\documentclass[10pt,twocolumn,letterpaper]{article}

\usepackage[pagenumbers]{cvpr} 

\usepackage[dvipsnames]{xcolor}

\usepackage{multirow}
\usepackage{makecell}
\usepackage{todonotes}
\usepackage{cuted}
\usepackage{tabularx}
\usepackage{chngcntr}
\usepackage{colortbl}
\usepackage{threeparttable}
\usepackage[accsupp]{axessibility} 
\usepackage{pifont}
\usepackage{filecontents,ifthen}
\usepackage{svg}
\usepackage{bm}
\usepackage{tcolorbox}

\definecolor{cvprblue}{rgb}{0.21,0.49,0.74}
\usepackage[pagebackref,breaklinks,colorlinks,allcolors=cvprblue]{hyperref}

\def\paperID{1529} 
\def\confName{CVPR}
\def\confYear{2025}

\title{
    PanSplat: 4K Panorama Synthesis with Feed-Forward Gaussian Splatting
}

\author{
    Cheng Zhang$^{1,2}$ \quad
    Haofei Xu$^{3}$ \quad
    Qianyi Wu$^{1}$\thanks{Corresponding author.}
    \\
    Camilo Cruz Gambardella$^{1,2}$ \quad
    Dinh Phung$^{1}$ \quad
    Jianfei Cai$^{1}$
    \\
    $^{1}$Monash University \quad
    $^{2}$Building 4.0 CRC, Caulfield East, Victoria, Australia \quad
    $^{3}$ETH Zurich
}

\begin{document}
\maketitle

\begin{strip}
    \vspace{-4.5em}
	\centering
	\small
    \newcommand{\rot}[1]{\rotatebox[origin=c]{90}{#1}}
    \newcommand{\outpano}[1]{\raisebox{-0.5\height}{\includegraphics[width=.49\linewidth,clip,trim=0 -40 0 0]{#1}}}
    \newcommand{\inpano}[1]{\raisebox{-0.5\height}{\includegraphics[width=.243\linewidth,clip,trim=0 -80 0 0]{#1}}}
    \setlength\tabcolsep{1px}
    \begin{tabular}{rccc}
        \rot{Output Novel View}
        & \outpano{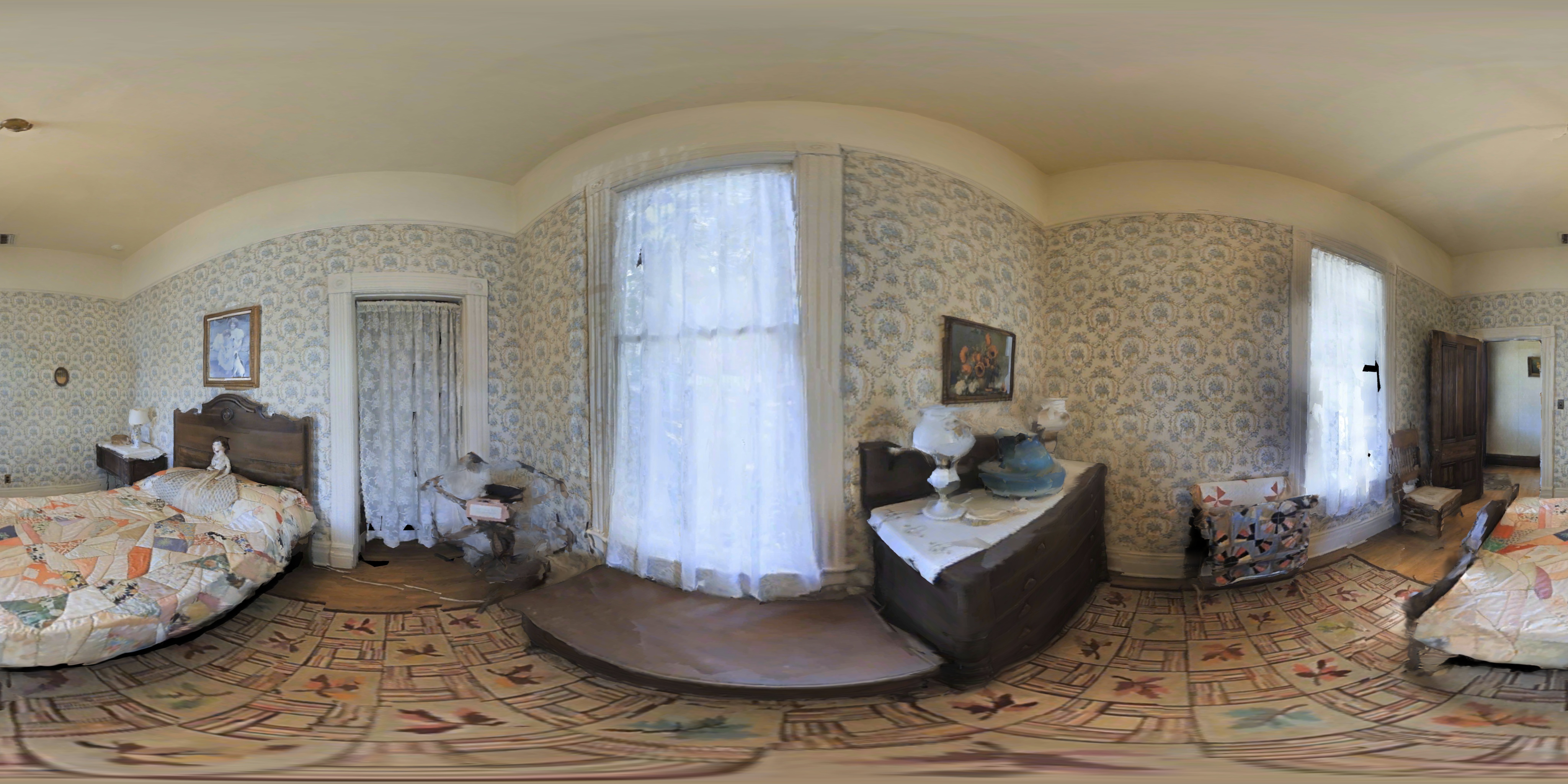}
        & \outpano{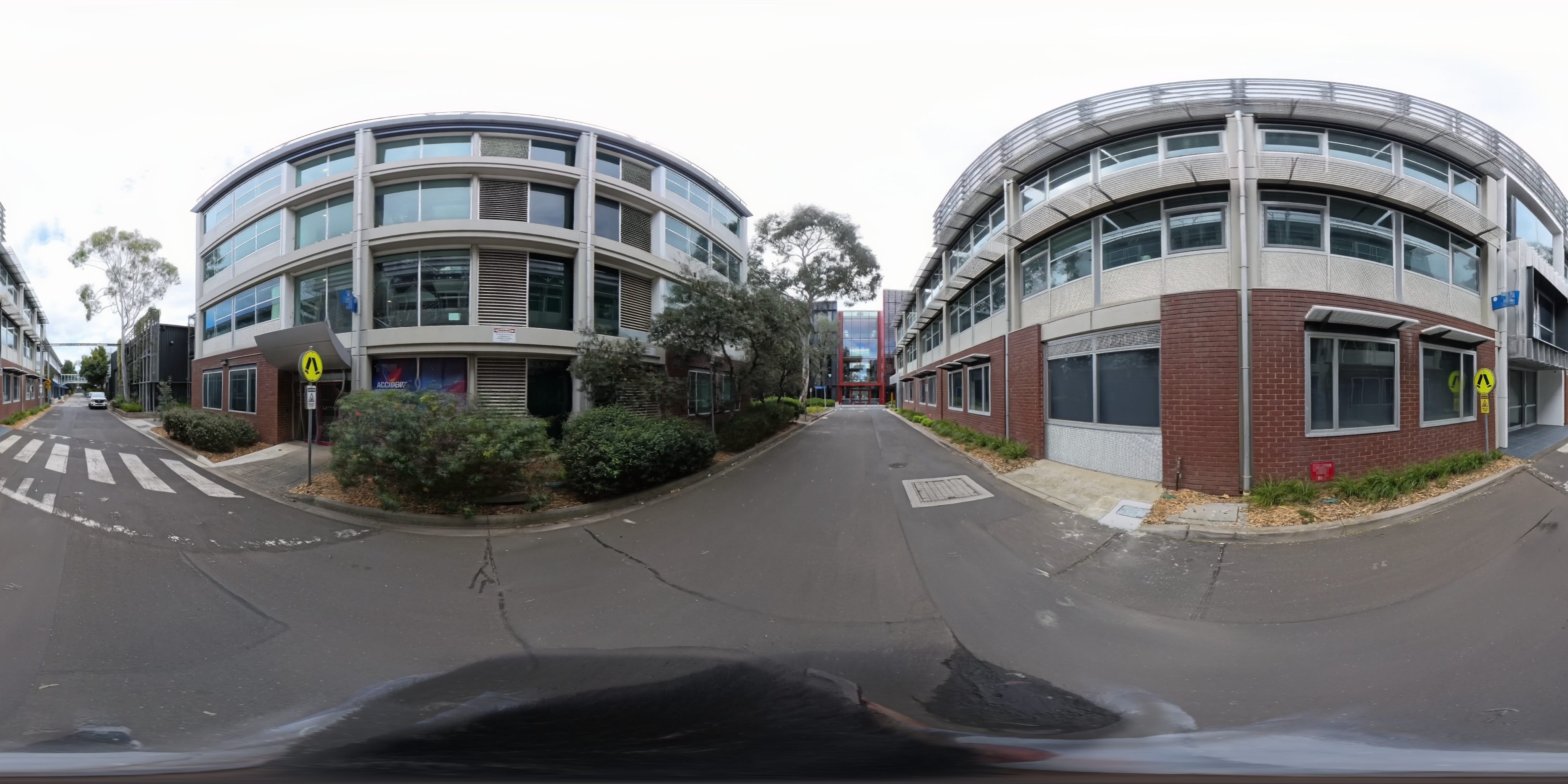}
        \\
        \rot{Input Views}
        & \makecell{
            \inpano{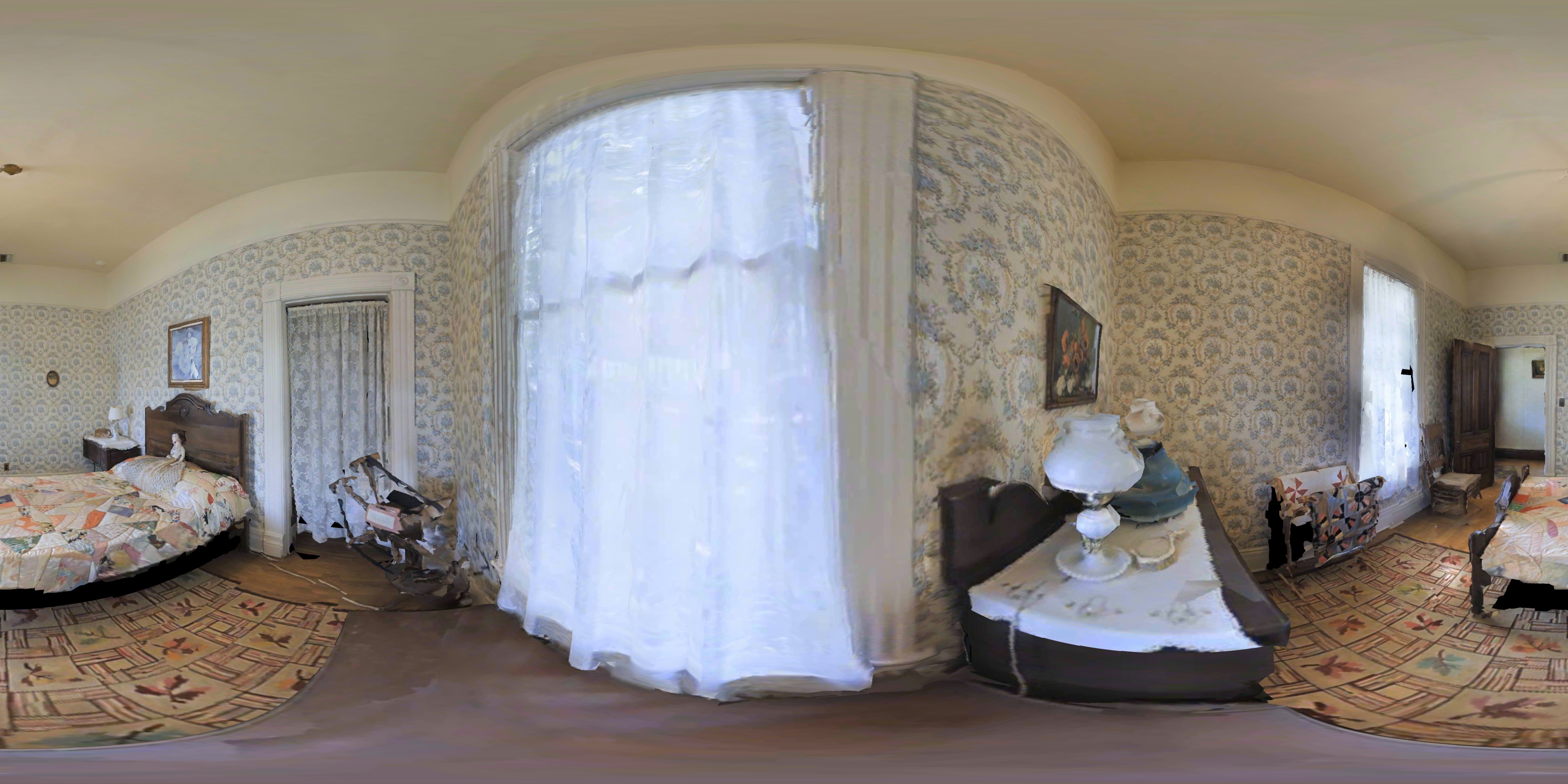}
            \inpano{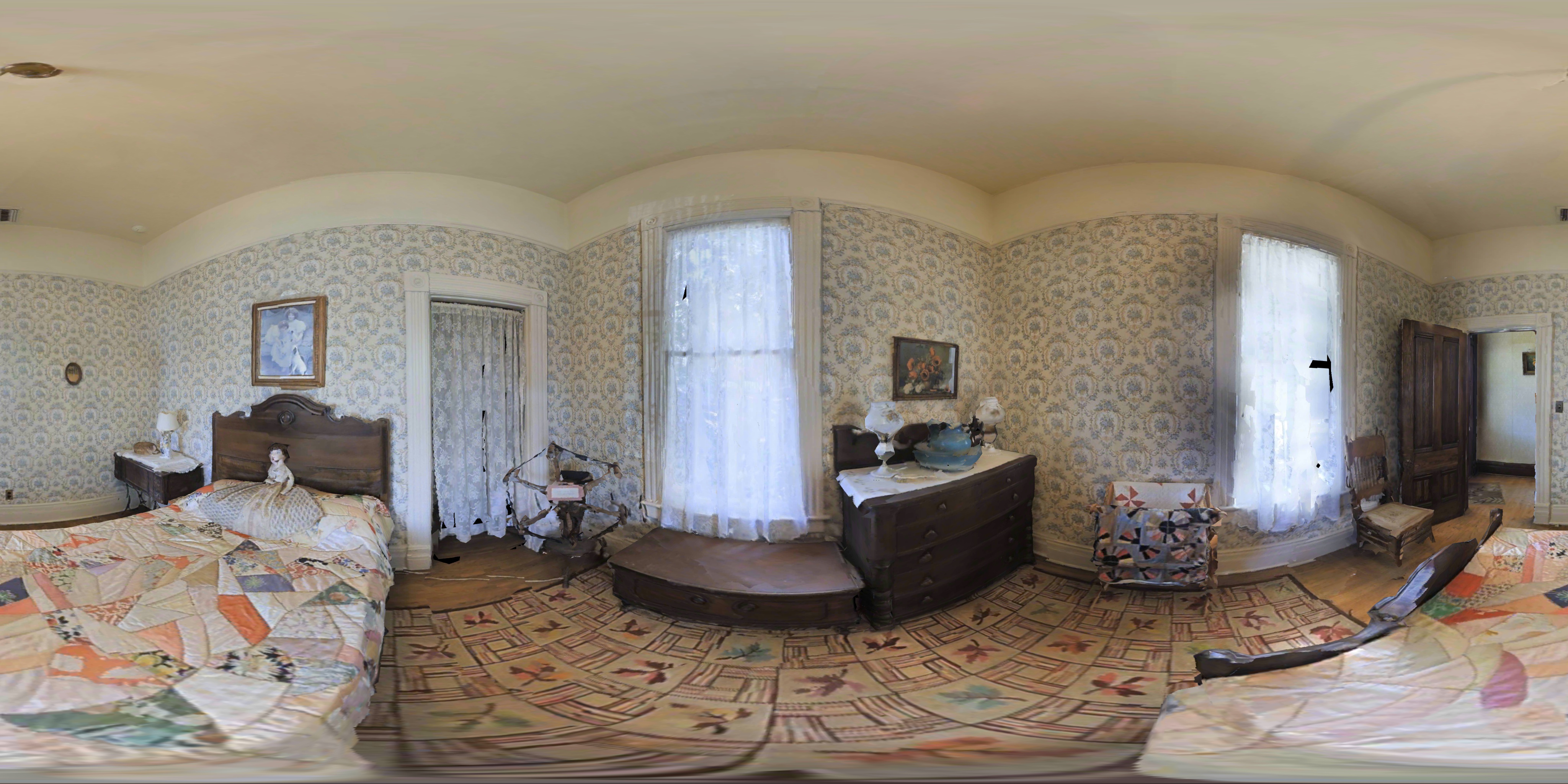}
        }
        & \makecell{
            \inpano{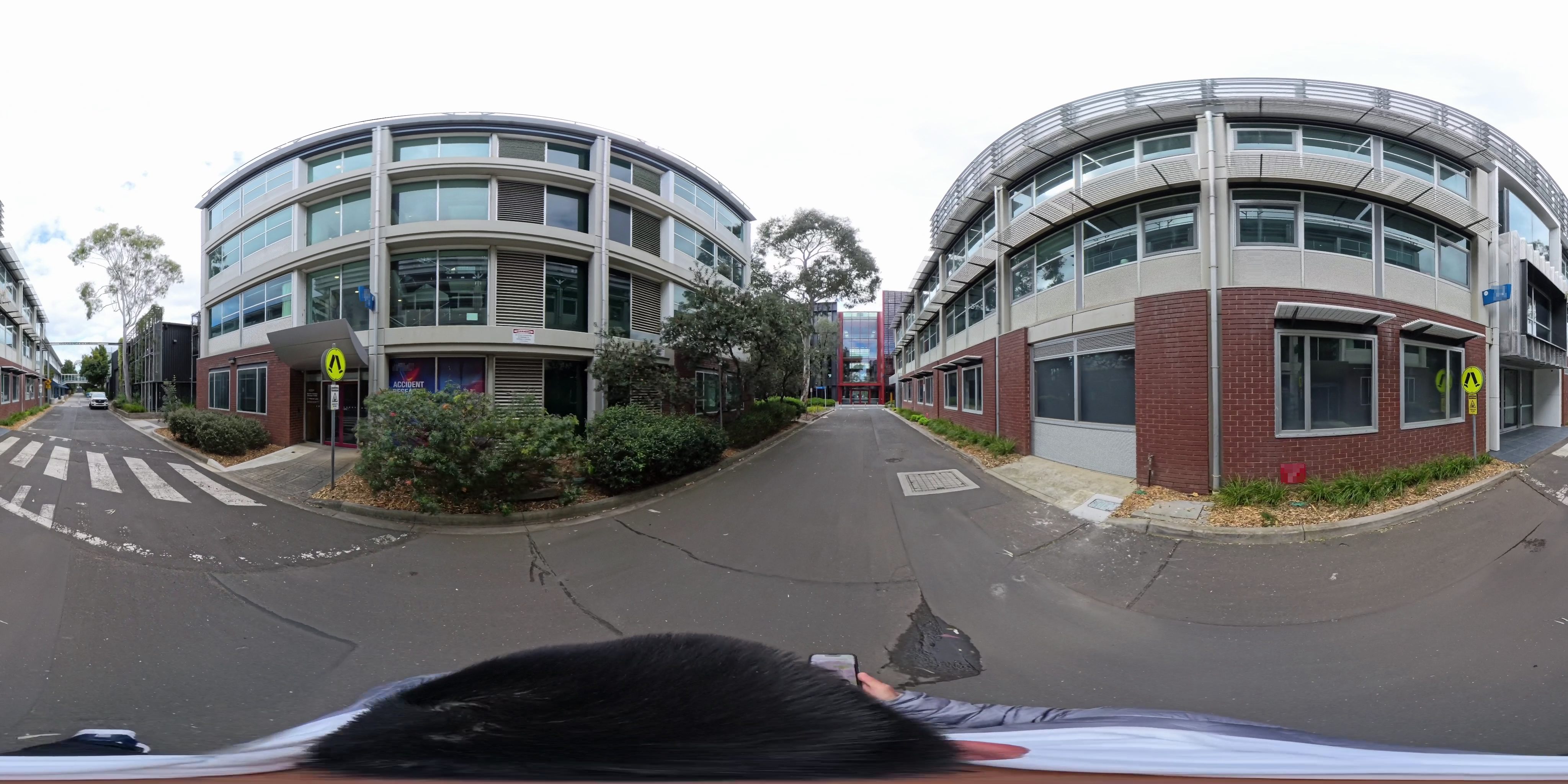}
            \inpano{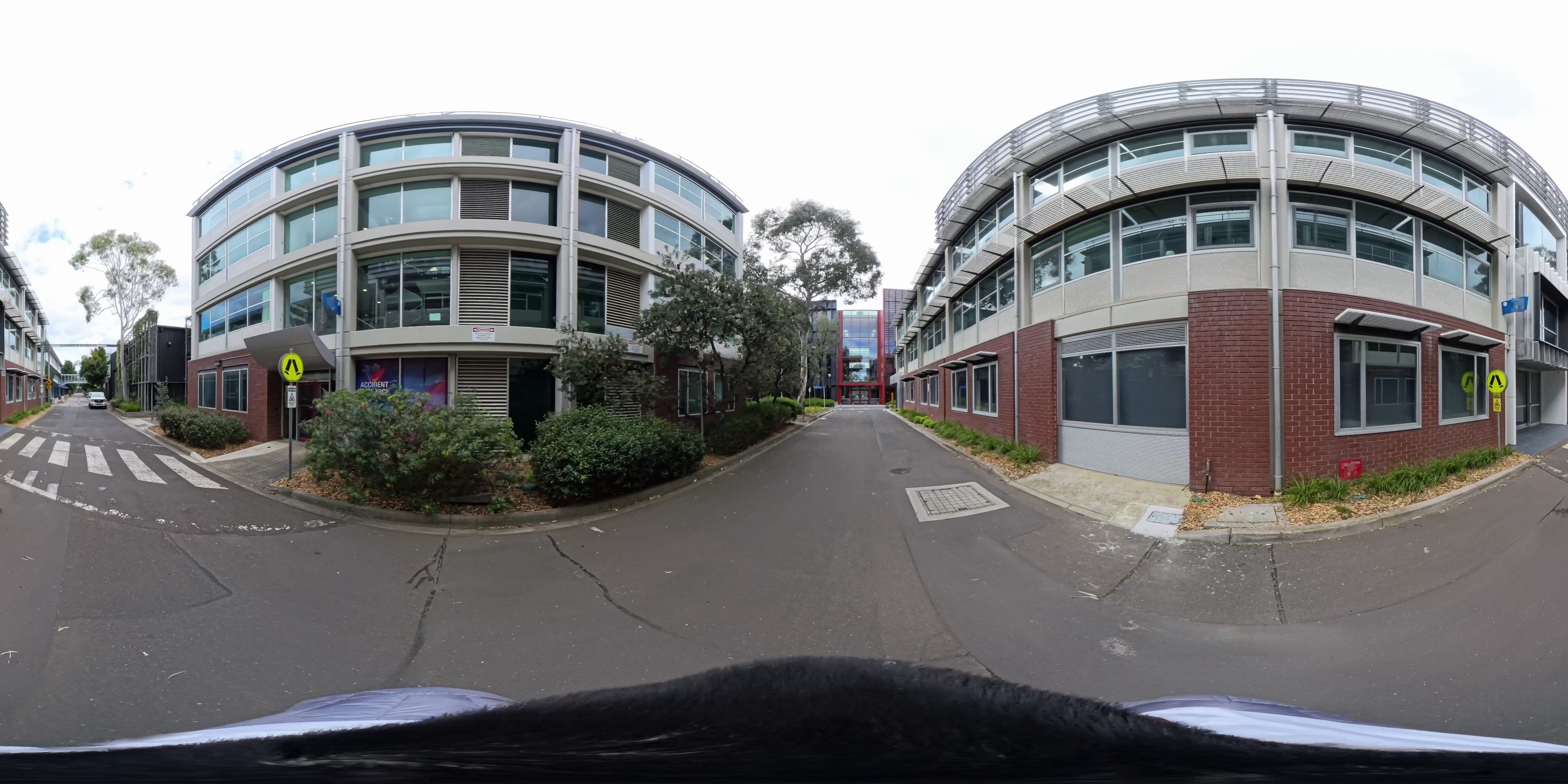}
        }
        \\
        & Trained on 4K panorama rendering of Matterport3D & Fine-tuned on real data and generalize to self-captured panoramas
    \end{tabular}
    \vspace{-0.5em}
	\captionof{figure}{
        \textbf{Our PanSplat can generate novel views from two 4K (2048 $\times$ 4096) panoramas.}
        We train on rendered Matterport3D~\cite{chang2017matterport3d} data at 4K resolution (left) and can generalize to 4K real-world data (right) with a few fine-tunings on 360Loc~\cite{huang2024360loc} data (Zoom in for details).
        Please refer to the supplementary video for more results.
    }\label{fig:teaser}
    \vspace{-1em}
\end{strip}

\begin{abstract}
    With the advent of portable 360° cameras, panorama has gained significant attention in applications like virtual reality (VR), virtual tours, robotics, and autonomous driving.
    As a result, wide-baseline panorama view synthesis has emerged as a vital task, where high resolution, fast inference, and memory efficiency are essential.
    Nevertheless, existing methods are typically constrained to lower resolutions ($512 \times 1024$) due to demanding memory and computational requirements.
    In this paper, we present \textbf{PanSplat}, a generalizable, feed-forward approach that efficiently supports \textbf{resolution up to 4K} ($2048 \times 4096$).
    Our approach features a tailored spherical 3D Gaussian pyramid with a Fibonacci lattice arrangement, enhancing image quality while reducing information redundancy.
    To accommodate the demands of high resolution, we propose a pipeline that integrates a hierarchical spherical cost volume and Gaussian heads with local operations, enabling two-step deferred backpropagation for memory-efficient training on a single A100 GPU.
    Experiments demonstrate that PanSplat achieves state-of-the-art results with superior efficiency and image quality across both synthetic and real-world datasets.
    Code is available at \url{https://github.com/chengzhag/PanSplat}.
\end{abstract}

\vspace{-1.0em}
\section{Introduction}\label{sec:intro}

The demand for rich visual content for virtual reality (VR) and virtual tours has surged alongside the rise of 360° cameras and immersive technologies.
Panoramic light field systems~\cite{overbeck2018system,broxton2020immersive} offer compelling solutions for delivering realistic, immersive experiences, by enabling users to explore environments from a range of arbitrary viewpoints within designated virtual spaces.
Recent advancements in 360° cameras simplify immersive content creation, driving applications like street view (Google Maps~\cite{googlemaps}, Apple Maps~\cite{applemaps}) and virtual tours (Matterport~\cite{matterport}, Theasys~\cite{theasys}), where novel view synthesis from wide-baseline panoramas is essential for providing smooth transitions between locations.

In recent years, deep learning has driven advancements in medical imaging~\cite{ronneberger2015unet,wu2024diversified} and robotics~\cite{cai2024neusis,li2021structdepth,li2023textslam,li2020textslam,li2025hier,li2024hi}, while also making significant progress in immersive content creation.
While current methods have extensively explored wide-baseline panorama view synthesis, they often struggle to balance computational efficiency, memory consumption, image quality, and resolution.
Conventional methods~\cite{habtegebrial2022somsi,attal2020matryodshka,li2021extending,li2022omnisyn} rely on explicit 3D scene representations such as Multi-Plane Images (MPI)~\cite{habtegebrial2022somsi,attal2020matryodshka,li2021extending} or mesh~\cite{li2022omnisyn}, which offer potential scalability to high resolutions but often yield lower image quality due to limited expressiveness.
Neural Radiance Fields (NeRF)-based methods~\cite{chen2023panogrf}, by contrast, deliver high-quality results but are computationally demanding and memory-intensive, making them less suitable for high-resolution panoramas.
Most existing methods are limited to a maximum resolution of $512 \times 1024$, which is well below 4K ($2048 \times 4096$), a resolution typically needed in VR applications for a truly immersive experience.

Recent trends in 3D Gaussian Splatting (3DGS)~\cite{kerbl20233d} have shown promising results in synthesizing novel views, marking a significant advancement in image quality and computational efficiency.
By representing scenes as collections of Gaussian primitives, 3DGS uses rasterization instead of volumetric sampling of NeRF to achieve high-quality, highly efficient rendering while also enabling differentiable rendering for training.
Subsequent works have further pushed the boundaries of 3DGS by introducing feed-forward networks~\cite{charatan2024pixelsplat,chen2025mvsplat} to predict Gaussians directly from input images, extending it to sparse view inputs.
Despite these advancements, existing 3DGS methods are not directly applicable to panoramas due to two primary challenges: 1) the unique spherical geometry of panoramas, which conflicts with pixel-aligned Gaussians and results in overlapping and redundant Gaussians near the poles; 2) the high-resolution demand of VR applications, which makes it infeasible for current methods (\eg, MVSplat~\cite{chen2025mvsplat}) to scale efficiently due to memory limitations.

\begin{figure}[t]
    \centering
    \includegraphics[width=0.98\linewidth, trim={0 20px 0 20px}, clip]{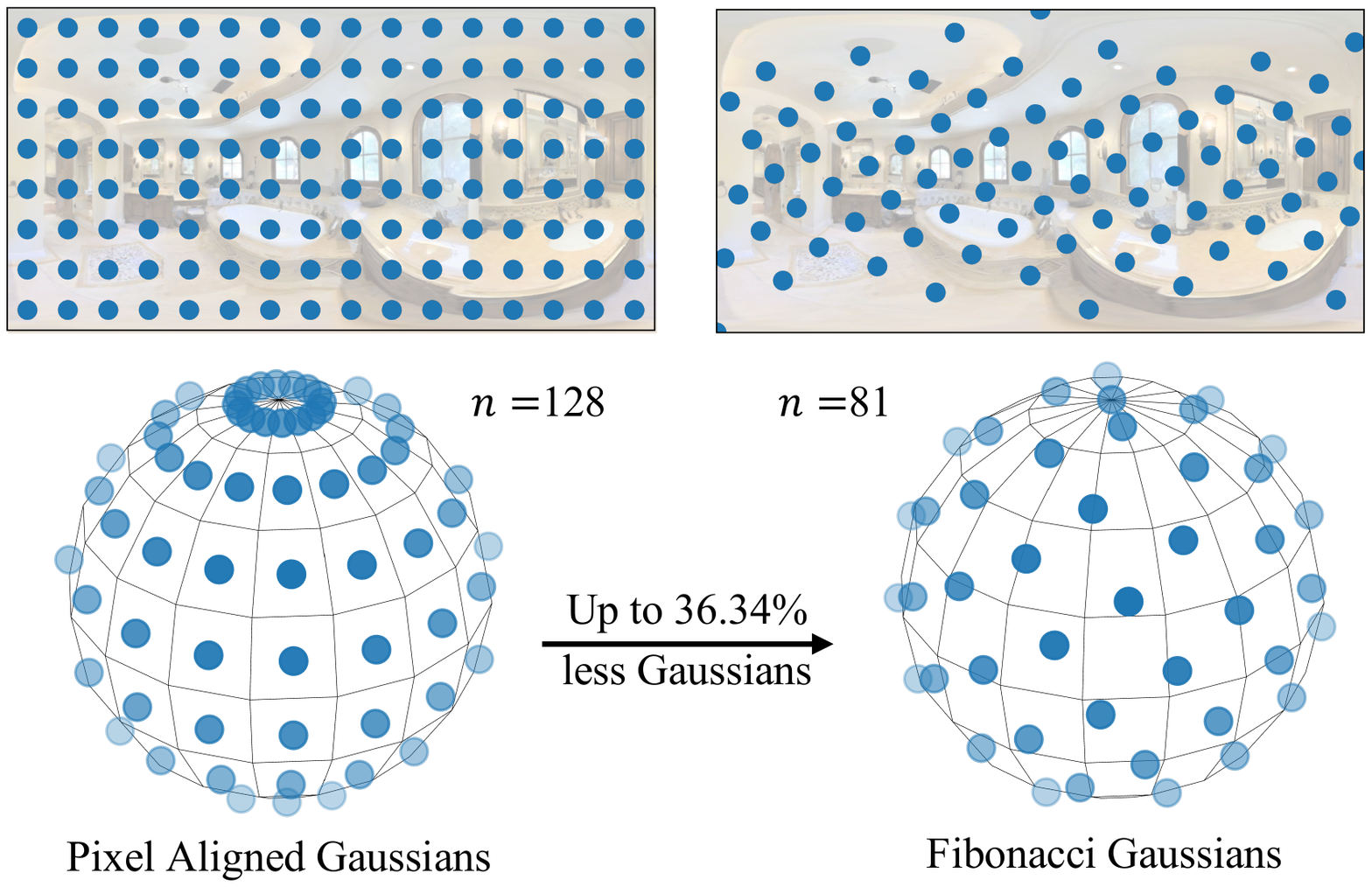}
    \vspace{-1em}
    \caption{
        \textbf{Fibonacci Gaussians.}
        We propose a Fibonacci lattice arrangement for the Gaussians to be distributed uniformly across the sphere, avoiding information redundancy near the poles, and significantly reducing the number of required Gaussians.
    }\label{fig:fibo}
    \vspace{-1em}
\end{figure}

In this work, we present \textbf{PanSplat}, a feed-forward approach optimized for 4K resolution inputs, generating a 3D Gaussian representation specifically tailored for panoramic formats to enable 4K novel view synthesis from wide-baseline panoramas (see examples in \cref{fig:teaser}).
To address the first challenge, we introduce a Fibonacci lattice arrangement for 3D Gaussians (illustrated in \cref{fig:fibo}), significantly reducing the required Gaussians by uniformly distributing them across the sphere. On the other hand, to enhance rendering quality, we implement 3D Gaussian pyramid, which represents the scene at multiple scales, capturing fine details across varying levels.
To address the second challenge, we utilize a hierarchical spherical cost volume built on a transformer-based network to estimate high-resolution 3D geometry with improved efficiency.
We then design Gaussian heads with local operations to predict Gaussian parameters, enabling two-step deferred backpropagation for memory-efficient training at 4K resolution.
Additionally, we introduce a deferred blending technique that reduces artifacts from misaligned Gaussians due to moving objects and depth inconsistencies, enhancing rendering quality in real-world scenes.

Our main contributions can be summarized as follows.
\begin{itemize}
    \item We present PanSplat, a feed-forward approach that efficiently generates high-quality novel views with spherical 3D Gaussian pyramid tailored for panorama formats.
    \item We design a pipeline featuring a hierarchical spherical cost volume and Gaussian heads with local operations, which enables a two-step deferred backpropagation, efficiently scaling to higher resolutions.
    \item We demonstrate that PanSplat achieves state-of-the-art results with superior image quality across synthetic and real-world datasets, with up to $\bm{70 \times}$ faster inference speed compared to the SOTA method~\cite{chen2023panogrf}. By supporting 4K resolution, PanSplat becomes a promising solution for immersive VR applications.
\end{itemize}

\begin{figure*}[t]
    \vspace{-1em}
    \centering
    \includegraphics[width=1.\linewidth, trim={0 0 0 0}, clip]{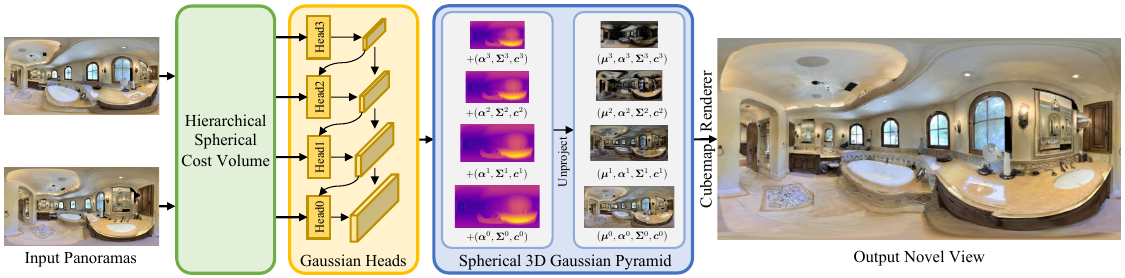}
    \vspace{-1.5em}
    \caption{
        \textbf{Our proposed PanSplat pipeline.}
        Given two wide-baseline panoramas, we first construct a hierarchical spherical cost volume (\cref{sec:cost_volume}) using a Transformer-based FPN to extract feature pyramid and 2D U-Nets to integrate monocular depth priors for cost volume refinement.
        We then build Gaussian heads (\cref{sec:gaussian_pred}) to generate a feature pyramid, which is later sampled with Fibonacci lattice and transformed to spherical 3D Gaussian pyramid (\cref{sec:s3dgp}).
        Finally, we unproject the Gaussian parameters for each level and view, consolidate them into a global representation, and splat it into novel views using a cubemap renderer.
        \emph{For simplicity, intermediate results of only a single view are shown.}
    }\label{fig:pipeline}
    \vspace{-1em}
\end{figure*}

\section{Related Work}\label{sec:related_work}

\noindent\textbf{Sparse Perspective Novel View Synthesis.}
The task of novel view synthesis has been widely explored for perspective images.
Recent methods such as NeRF~\cite{mildenhall2020nerf} and 3DGS~\cite{kerbl20233d} have achieved remarkable results but rely heavily on dense input views, making them costly for real-world applications.
To address this limitation, many approaches have emerged that leverage prior knowledge from large-scale datasets to handle sparse input views.
These include per-scene optimization methods~\cite{niemeyer2022regnerf,truong2023sparf,wu2024reconfusion,deng2022depth,yu2022monosdf} that optimize a scene-specific model, as well as feed-forward methods that directly predict novel views from sparse inputs~\cite{chen2021mvsnerf,chen2023explicit,xu2024murf,charatan2024pixelsplat,chen2025mvsplat,wewer2024latentsplat,zhang2025gs,li2024GGRt,tang2024hisplat,ye2024no,liu2022neural,wang2021ibrnet,smart2024splatt3r} or single view~\cite{szymanowicz2024splatter,szymanowicz2024flash3d}.
While these methods simplify data capture requirements, optimization-based approaches remain computationally expensive and require significant time to train a model specific to each scene.
Feed-forward methods like NeuRay~\cite{liu2022neural}, IBRNet~\cite{wang2021ibrnet}, and MVSplat~\cite{chen2025mvsplat}, on the other hand, are more efficient by utilizing pre-trained models that generalize well across different scenes.
Despite recent advancements, these methods are not directly applicable to panoramas due to their distinct spherical geometry.
Our approach builds upon the feed-forward 3DGS framework, extending it to high-resolution panoramas by introducing a tailored spherical 3D Gaussian pyramid and a hierarchical spherical cost volume.
While concurrent work~\cite{tang2024hisplat} also explores hierarchical 3D Gaussians as a more expressive representation, it does not address the unique challenges of high-resolution or panoramic formats.

\noindent\textbf{Sparse Panorama Novel View Synthesis.}
Recently, the panorama format has gained significant attention as it becomes more accessible and valuable in applications like VR, virtual tours, and autonomous driving, with numerous works focusing on generation~\cite{tang2023mvdiffusion,hollein2023text2room,zhang2024taming,ye2024diffpano}, out-painting~\cite{wang2022stylelight,akimoto2022diverse,wang2023360,oh2022bips,wu2023ipo,dastjerdi2022guided}, and reconstruction~\cite{kim2024omnisdf,jang2022egocentric,zhang2021deeppanocontext,dong2024panocontext,yang2022neural}.
However, novel view synthesis for panoramas has received less attention compared to perspective images, largely due to the challenges in efficiently processing high-resolution equirectangular images.
Existing methods~\cite{choi2024omnilocalrf,li2024omnigs,chen2022casual,bai2024360,huang2022360roam,choi2023balanced} have focused on per-scene optimization, addressing the distinct spherical geometry of panoramas.
Further advancements have been made for sparse panorama inputs, such as SOMSI~\cite{habtegebrial2022somsi}, which takes a set of panorama images and represents 3D scene with a Multi-Sphere Images (MSI) representation.
OmniSyn~\cite{li2022omnisyn} further reduces the input requirement to two wide-baseline panoramas, but the less expressive mesh representation limits the quality of novel views.
Following this setting, PanoGRF~\cite{chen2023panogrf} enhances image quality with a spherical NeRF and combines a monocular and stereo depth model to improve geometry, but is computationally expensive due to volumetric sampling of NeRF.
Concurrent work~\cite{chen2024splatter} also explores 3DGS for panoramas, but it does not address the unique challenges of high-resolution on real-world datasets.
In contrast, our PanSplat is designed to efficiently handle high-resolution panoramas, capable of providing a realistic and immersive experience.

\section{Method}\label{sec:method}

PanSplat is a feed-forward model that synthesizes high-quality novel views efficiently from two posed wide-baseline panoramas as shown in \cref{fig:pipeline}.
We introduce a spherical 3D Gaussian pyramid (\cref{sec:s3dgp}) specifically tailored for panoramic images, which we pair with a hierarchical spherical cost volume (\cref{sec:cost_volume}) and Gaussian heads (\cref{sec:gaussian_pred}) to enable scalable, high-resolution output up to 4K for real-world applications.
The training procedure is described in detail in \cref{sec:training}.

\subsection{Spherical 3D Gaussian Pyramid}\label{sec:s3dgp}

\noindent\textbf{Fibonacci Gaussians.}
Recall that current pixel-aligned Gaussian splatting methods~\cite{charatan2024pixelsplat,chen2025mvsplat,zhang2025gs} assign a Gaussian to each pixel (top-left of \cref{fig:fibo}), where each Gaussian is defined by parameters including center $\bm{\mu}$, opacity ${\alpha}$, covariance $\bm{\Sigma}$, and color $\bm{c}$.
Such representation is inefficient for panoramas, as pixel density varies with latitude, leading to redundant Gaussians near the poles, as shown in the bottom-left of \cref{fig:fibo}.
Instead, we propose to distribute the Gaussians using a Fibonacci lattice~\cite{roberts2020how,fibonacci,frisch2021deterministic} to achieve a more uniform distribution across the sphere (bottom-right of \cref{fig:fibo}), which significantly reduces Gaussian redundancy, particularly near the poles (top-right of \cref{fig:fibo}).
Specifically, we set the number of Gaussian ${n} = \lfloor {W}^2 / \pi \rfloor$ based on image resolution, where $W$ is the panorama image width, to ensure Gaussian density near the equator is similar to that of image pixels.
The value of $n$ can be adjusted to balance image quality and rendering efficiency.
Then, for the $j$-th Gaussian on the Fibonacci lattice, its coordinates on the image plane are calculated as $({x}_{j}, {y}_{j}) = \left( \frac{j}{\phi} \mod 1, \frac{j}{{n} - 1} \right)$, where $\phi = \frac{1 + \sqrt{5}}{2}$ is the golden ratio.
%
This configuration reduces Gaussian usage by up to 36.34\% compared to pixel-aligned splatting without compromising image qualilty (see +Fibo in \cref{tbl:ablation}).

\noindent\textbf{3D Gaussian Pyramid.}
To further enhance image quality, we draw inspiration from~\cite{hyun2024adversarial} to introduce a pyramid structure that captures multi-scale details.
Given two input panoramas $\{\bm{I}_{i}\}_{i = 0}^{1} \in \mathbb{R}^{H \times W \times 3}$, we aim to predict Gaussian parameters $\{( \bm{\mu}^{l}_{i}, \bm{\alpha}^{l}_{i}, \bm{\Sigma}^{l}_{i}, \bm{c}^{l}_{i} )\}_{l=0, i=0}^{L-1, 1}$ at different levels $l$ for each view $i$.
To estimate the Gaussian centers $\bm{\mu}$, we first predict a depth for each Gaussian and then unproject the image-plane coordinates $({x}_{j}, {y}_{j})$ into 3D space.
We define the number of Gaussians at level $l$ as ${n}^{l} = \lfloor {W}^{2} / (2^{l} \pi) \rfloor$, with the number of pyramid levels set to ${L} = 4$.
Each level is designed to represent a specific level of details, ranging from the coarsest level, $l = 3$, with the fewest Gaussians, to the finest level, $l = 0$, which has the highest Gaussian density.

\subsection{Hierarchical Spherical Cost Volume}\label{sec:cost_volume}

To support the proposed pyramid representation and meet the high-resolution demands of real-world applications, we construct a hierarchical spherical cost volume that efficiently estimates 3D geometry at a higher resolution than MVSplat~\cite{chen2025mvsplat}.

\noindent\textbf{Feature Pyramid Extraction.}
We first apply a Feature Pyramid Network (FPN)~\cite{lin2017feature} to extract multi-scale features from the input panoramas $\{\bm{I}_{i}\}_{{i} = 0}^{1}$.
At the coarsest level of the FPN, we introduce a Swin Transformer~\cite{liu2021swin} with cross-view attention, enabling effective information exchange between the two panoramas for improved matching.
We denote the image feature pyramid as $\{\bm{F}^{l}_{i}\}_{{l} = 0}^{{L} - 1} \in \mathbb{R}^{{H} / 2^{l} \times {W} / 2^{l} \times {C}^{l}}$, where ${C}^{l}$ represents the number of channels at level ${l}$.
The feature pyramid is designed to match the ${L}$ levels of the Gaussian pyramid, serving as an additional input for predicting Gaussian parameters in \cref{sec:gaussian_pred}.

\noindent\textbf{Spherical Cost Volume Initialization.}
Building on this feature representation, we proceed to construct a hierarchical cost volume~\cite{yang2020cost,gu2020cascade}, beginning at the coarsest level ${l} = 3$.
For each reference view ${i} = 0,1$, we uniformly sample ${D}$ inverse depth candidates within a preset range $[d_{\text{min}}, d_{\text{max}}]$ and warp the coarsest feature maps $\bm{F}^{3}_{1 - {i}}$ to the corresponding reference view using spherical projection~\cite{li2022omnisyn,chen2023panogrf}.
We then calculate the correlations to reference features $\bm{F}^{3}_{i}$ via a dot product~\cite{xu2023unifying}, resulting in a cost volume $\bm{C}^{3}_{i} \in \mathbb{R}^{{H} / 8 \times {W} / 8 \times {D}}$ for each view.
To regularize the cost volume in occluded or texture-less regions, we integrate pre-trained monocular depth features~\cite{jiang2021unifuse}.
Specifically, a 2D U-Net~\cite{ronneberger2015unet} takes in the concatenated monocular depth features, cost volume, and reference features, and produces a residual that refines the cost volume.
The refined cost volume $\bm{\tilde{C}}^{3}_{i}$ is then normalized with $\texttt{softmax}$ along the depth dimension, yielding the probability distribution of object surfaces across different depths, which we use to weight and average the depth candidates, resulting in the initial depth prediction $\bm{D}^{3}_{i}$.

\noindent\textbf{Hierarchical Cost Volume Refinement.}
We refine the depth predictions at progressively finer levels $l = 2, 1$, where each level searches near the coarse depth from the previous level and generates a higher-resolution cost volume.
Specifically, we up-sample $\bm{D}^{l + 1}_{i}$ to the next level $l$, then construct a more compact cost volume with ${D} / 2^{3 - l}$ depth candidates within a reduced range $({d}_{\text{max}} - {d}_{\text{min}}) / 2^{3 - l}$.
Independent 2D U-Net for each level is then used to refine the cost volume, with an additional input $\bm{D}^{l + 1}_{i}$ to provide contextual information.
This process ultimately yields a cost volume $\bm{\tilde{C}}^{1}_{i} \in \mathbb{R}^{{H} / 2 \times {W} / 2 \times {D} / 4}$ for each view, along with depth predictions $\{\bm{D}^{l}_{i}\}^{3}_{{l} = 1}$ across different levels.
To balance memory consumption with depth accuracy, we skip refinement at the finest level ${l} = 0$, achieving $2 \times$ depth resolution compared to MVSplat under a similar memory budget.

\subsection{Gaussian Parameter Prediction and Rendering}\label{sec:gaussian_pred}

\noindent\textbf{Gaussian Heads.}
After constructing the hierarchical cost volume, we design light-weight Gaussian heads to predict Gaussian parameters at different levels for each view.
At level $l$, we resize the cost volume $\bm{\tilde{C}}^{1}_{i}$ and the input image $\bm{I}_{i}$ to match the resolution of the image feature $\bm{F}^{l}_{i}$, then concatenate them as input.
Since we define Gaussians on a Fibonacci lattice, we do not predict the Gaussian parameters in a pixel-aligned manner.
Instead, for each level, we use a CNN to first extract a feature map $\bm{\tilde{F}}^{l}_{i}$, then interpolate a feature vector for each Gaussian, followed by a fully connected layer to predict the Gaussian parameters.
So far, we assume that the different layers of Gaussians can represent different levels of details in the scene to improve the rendering quality.
However, we find that the network does not fully utilize the multi-scale information (\cref*{sec:more_ablation} in supplementary material), which is likely due to the lack of guidance between different levels.
Therefore, we introduce a residual design by up-sampling the feature map $\bm{\tilde{F}}^{l + 1}_{i}$ from the previous level, concatenating it as an additional input to the current level Gaussian head, and predicting a residual based on this feature map.
This design functions as a skip connection, enforcing dependencies between adjacent levels and guiding the network to leverage the multi-scale structure in a coarse-to-fine manner.

\noindent\textbf{Cubemap Renderer.}
We consolidate the Gaussian parameters from two input views and different levels of Gaussian heads to form a single Gaussian representation, which we then render in novel views using a cubemap renderer.
Specifically, we first render 6 cameras with $90^{\circ}$ field of view (FOV) at the same position but facing different directions defined by the cubemap faces.
Then we stitch the cubemap into a panorama with differentiable grid sampling operation (see \cref*{sec:defbp} in the supplementary material for details).
Although existing methods~\cite{li2024omnigs,bai2024360} provide renderers with improved splatting accuracy for panoramas, they are not designed for memory efficiency.
In contrast, we reduce memory consumption for high-resolution training by integrating the cubemap renderer and the Gaussian heads with a two-step deferred backpropagation approach.

\noindent\textbf{Two-step Deferred Backpropagation.}
Based on the observation that image quality relies more on texture resolution than on geometry resolution, we leverage the decoupled design of geometry (hierarchical cost volume) and appearance (Gaussian heads) to scale efficiently to higher resolutions.
Specifically, we down-sample the input image for the hierarchical cost volume to $512 \times 1024$ while preserving the input resolution for the Gaussian heads.
Between the two modules, image features and cost volumes from the former are up-sampled to match the resolution of the latter.
This approach allows the finest level of Gaussians to be predicted using full-resolution images as input, preserving detailed texture information, while the geometry is estimated at a lower resolution to maintain reasonable memory usage.
Although this design significantly reduces memory consumption (see PanSplat in \cref{fig:gpu_memory}), it still falls short of handling 4K resolution due to the considerable memory demands of both the Gaussian heads and the Gaussian renderer.
For inference, we exploit the local operations of Gaussian heads to enable tiled operations, while the cubemap renderer supports sequential face rendering, both contributing to enhanced memory efficiency.
Inspired by~\cite{zhang2022arf,li2024GGRt}, we further design a two-step deferred backpropagation to enable memory-efficient training at 4K resolution.
In this approach, we first disable auto-differentiation to render the full panorama, compute the image loss, and cache gradients on the image.
Subsequently, we enable auto-differentiation and backpropagate gradients in a ``two-step'' manner: first, the panorama is re-rendered face by face, backpropagating and accumulating gradients to the Gaussian parameters; second, the Gaussian parameters are re-generated tile by tile, with gradients backpropagated and accumulated to the network parameters.

\noindent\textbf{Deferred blending.}
Due to the omnidirectional nature of panoramas, it is inevitable to include moving objects when capturing real datasets, \eg, camera operators, pedestrians, or vehicles.
In this scenario, the two input views would produce inconsistent depth and misaligned Gaussians, leading to artifacts in the rendered images.
To mitigate this issue, we draw inspiration from~\cite{wang2021ibrnet} and introduce a deferred blending approach.
For details, please refer to \cref*{sec:ext_to_real} in the supplementary material.

\subsection{Training}\label{sec:training}

\noindent\textbf{Synthetic Data.}
We follow PanoGRF~\cite{chen2023panogrf} to train PanSplat on synthetic data with depth and image losses.
For depth supervision, we use $L_1$ loss on the depth predictions from three-level hierarchical cost volume:
\begin{equation}
    \mathcal{L}_{\text{depth}} = \sum_{{i} = 0,1} \sum_{{l} = 1}^{3} {\gamma}^{{l} - 1} \left\| \bm{D}^{l}_{i} - \bm{\hat{D}}^{l}_{i} \right\|_1,
\end{equation}
where ${\bm{\hat{D}}^{l}_{i}}$ denotes down-sampled ground truth depth, and ${\gamma}$ is a weight that emphasizes finer levels. 
For image supervision, we compute $L_2$ and LPIPS~\cite{zhang2018unreasonable} losses between the rendered image $\bm{I}$ and the ground truth image $\bm{\hat{I}}$:
\begin{equation}
    \mathcal{L}_{\text{rgb}} = \left\| \bm{I} - \bm{\hat{I}} \right\|_2 + \lambda \text{LPIPS}(\bm{I}, \bm{\hat{I}}),
\end{equation}
We jointly train the network using loss function $\mathcal{L}_{\text{synthetic}} = {\alpha} \mathcal{L}_{\text{depth}} + \mathcal{L}_{\text{rgb}}$, with ${\gamma} = 0.9$, ${\lambda} = 0.1$ and ${\alpha} = 0.05$.

\noindent\textbf{Real Data.}
It is challenging to obtain ground truth depth for real-world data.
Fortunately, recent works~\cite{chen2025mvsplat,wewer2024latentsplat} demonstrate that depth estimation can be learned with a self-supervised approach using Gaussian splatting. 
Since all levels of the hierarchical cost volume require supervision, we propose adding auxiliary Gaussian heads to each level, replacing direct depth loss for training purposes. 
These auxiliary Gaussian heads operate in parallel with the main Gaussian heads in the network and do not share the same residual design. 
To enable direct gradient flow, we directly use the predicted depth from the cost volume at each level to unproject Gaussian centers.
For simplicity, only 2 CNN layers are used to predict the other Gaussian parameters.
The predicted Gaussians from each level are then separately rendered to panoramas $\{\bm{I}^{l}\}^{3}_{l = 1}$ and compared with the ground truth using image loss $\mathcal{L}_{\text{rgb}}$.
The final loss function becomes
\begin{equation}
    \mathcal{L}_{\text{real}} = \sum_{{l} = 1}^{3} {\gamma}^{{l} - 1} \mathcal{L}_{\text{rgb}} (\bm{I}^{l}, \bm{\hat{I}}) + \mathcal{L}_{\text{rgb}} (\bm{I}, \bm{\hat{I}}).
\end{equation}

\section{Experiment}\label{sec:exp}

\begin{table*}[t]
    \setlength\tabcolsep{1pt}
    \centering
    \resizebox*{1.0\linewidth}{!}{
    \begin{tabular}{cccccccccccccccc}
        \toprule
        Dataset & \multicolumn{9}{c}{Matterport3D} & \multicolumn{3}{c}{Replica} & \multicolumn{3}{c}{Residential} \\
        \cmidrule(lr){1-1}\cmidrule(lr){2-10}\cmidrule(r){11-13}\cmidrule(r){14-16}
        Baseline  & \multicolumn{3}{c}{1.0m} & \multicolumn{3}{c}{1.5m} & \multicolumn{3}{c}{2.0m} & \multicolumn{3}{c}{1.0m} & \multicolumn{3}{c}{about 0.3m} \\
        \cmidrule(lr){1-1}\cmidrule(lr){2-4}\cmidrule(lr){5-7}\cmidrule(lr){8-10}\cmidrule(r){11-13}\cmidrule(r){14-16}
        Method & WS-PSNR$\uparrow$ & SSIM$\uparrow$ & LPIPS$\downarrow$ & WS-PSNR$\uparrow$ & SSIM$\uparrow$ & LPIPS$\downarrow$ & WS-PSNR$\uparrow$ & SSIM$\uparrow$ & LPIPS$\downarrow$ & WS-PSNR$\uparrow$ & SSIM$\uparrow$ & LPIPS$\downarrow$ & WS-PSNR$\uparrow$ & SSIM$\uparrow$ & LPIPS$\downarrow$ \\
        \cmidrule(lr){1-1}\cmidrule(lr){2-4}\cmidrule(lr){5-7}\cmidrule(lr){8-10}\cmidrule(r){11-13}\cmidrule(r){14-16}
        S-NeRF~\cite{mildenhall2020nerf}  & 15.25 & 0.579 & 0.546 & 14.16 & 0.563 & 0.580 & 13.13 & 0.523 & 0.607 & 16.10 &0.723 & 0.443 & 22.47 & 0.741 & 0.435 \\
        OmniSyn~\cite{li2022omnisyn} & 22.90 & 0.850 & 0.244 & 20.31 & 0.790 & 0.317 & 18.91 & \cellcolor{orange!30}0.761 & \cellcolor{yellow!30}0.354 & 23.17 & 0.898&0.189&-&-&-\\
        IBRNet~\cite{wang2021ibrnet}  & 25.72 & 0.855 & 0.258 & 21.69 & 0.751 & 0.382 & \cellcolor{yellow!30}20.04 & 0.706 & 0.431 & 22.65 & 0.854 & 0.291 & 22.47 & 0.735 &0.498 \\
        NeuRay~\cite{liu2022neural} & 24.92 & 0.832 & 0.260 & 21.92 & 0.766 & 0.347 & 19.85 & \cellcolor{yellow!30}0.715 & 0.407 & 25.90 & 0.899 & 0.187 & 22.38 & 0.753 &0.427 \\
        PanoGRF~\cite{chen2023panogrf} & \cellcolor{yellow!30}27.12 & \cellcolor{yellow!30}0.876 & \cellcolor{yellow!30}0.195 & \cellcolor{orange!30}23.38 & \cellcolor{orange!30}0.811 & \cellcolor{yellow!30}0.282 & \cellcolor{red!30}20.96 & \cellcolor{orange!30}0.761 & \cellcolor{orange!30}0.352 & \cellcolor{yellow!30}29.22 & \cellcolor{yellow!30}0.937 & \cellcolor{yellow!30}0.134 & \cellcolor{orange!30}31.03 & \cellcolor{orange!30}0.909 & \cellcolor{yellow!30}0.207 \\
        MVSplat~\cite{chen2025mvsplat} & \cellcolor{orange!30}28.19 & \cellcolor{orange!30}0.912 & \cellcolor{orange!30}0.105 & \cellcolor{yellow!30}21.82 & \cellcolor{yellow!30}0.807 & \cellcolor{orange!30}0.230 & 13.31 & 0.595 & 0.554 & \cellcolor{orange!30}30.54 & \cellcolor{orange!30}0.958 & \cellcolor{red!30}0.059 & \cellcolor{red!30}31.21 & \cellcolor{yellow!30}0.906 & \cellcolor{orange!30}0.200 \\ 
        PanSplat & \cellcolor{red!30}28.81 & \cellcolor{red!30}0.931 & \cellcolor{red!30}0.091 & \cellcolor{red!30}24.09 & \cellcolor{red!30}0.849 & \cellcolor{red!30}0.181 & \cellcolor{orange!30}20.56 & \cellcolor{red!30}0.777 & \cellcolor{red!30}0.265 & \cellcolor{red!30}30.78 & \cellcolor{red!30}0.962 & \cellcolor{orange!30}0.069 & \cellcolor{yellow!30}30.97 & \cellcolor{red!30}0.917 & \cellcolor{red!30}0.172 \\ 
        \bottomrule
    \end{tabular}
    }
    \vspace{-0.5em}
    \caption{
        \textbf{Quantitative comparisons on synthetic datasets.}
        All models are trained on Matterport3D with a baseline of 1.0 meter and evaluated on the test set with the same baseline, as well as on wider baselines of 1.5 and 2.0 meters.
        Additionally, we evaluate on the Replica and Residential datasets to assess generalization to unseen data.
        Top results are highlighted in \colorbox{red!30}{top1}, \colorbox{orange!30}{top2}, and \colorbox{yellow!30}{top3}.
    }\label{tbl:cmp_syn}
\end{table*}

\begin{table*}[t]
    \setlength\tabcolsep{4pt}
    \centering
    \begin{tabular}{ccccccccc}
        \toprule
        Dataset & \multicolumn{4}{c}{360Loc (avg. 1.40m baseline)}&\multicolumn{4}{c}{Insta360 (16 frames apart)}\\
        \cmidrule(r){1-1}    \cmidrule(r){2-5} \cmidrule(r){6-9}
        Method & PSNR$\uparrow$ & WS-PSNR$\uparrow$ & SSIM$\uparrow$ & LPIPS$\downarrow$ & PSNR$\uparrow$ & WS-PSNR$\uparrow$ & SSIM$\uparrow$ & LPIPS$\downarrow$\\
        \cmidrule(r){1-1}    \cmidrule(r){2-5} \cmidrule(r){6-9}
        MVSplat~\cite{chen2025mvsplat} & 24.13 & 24.67 & 0.823 & 0.170 & 20.93 & 23.24 & 0.786 & 0.227 \\ 
        PanSplat & \cellcolor{orange!30}24.96 & \cellcolor{orange!30}25.58 & \cellcolor{orange!30}0.833 & \cellcolor{orange!30}0.159 & \cellcolor{orange!30}21.92 & \cellcolor{orange!30}24.43 & \cellcolor{orange!30}0.813 & \cellcolor{orange!30}0.211 \\ 
        PanSplat (w/ Deferred BL) & \cellcolor{red!30}27.35 & \cellcolor{red!30}28.14 & \cellcolor{red!30}0.860 & \cellcolor{red!30}0.127 & \cellcolor{red!30}23.36 & \cellcolor{red!30}25.68 & \cellcolor{red!30}0.822 & \cellcolor{red!30}0.183 \\ 
        \bottomrule
    \end{tabular}
    \vspace{-0.5em}
    \caption{
        \textbf{Quantitative comparisons on real-world datasets.}
        We compare with MVSplat, as it does not require depth supervision, which is unavailable for real-world datasets.
        All models are fine-tuned on the 360Loc dataset and directly tested on the Insta360 dataset for generalization evaluation.
    }\label{tbl:cmp_real}
    \vspace{-1em}
\end{table*}

{
\newcommand{\cmponerow}[1]{
    \begin{figure*}[t]
        \centering
        \small
        \setlength\tabcolsep{1px}
        \resizebox{1.\linewidth}{!}{
            \begin{tabular}{rcccccc}
                #1
                & Input Views & GT View & PanoGRF & MVSplat & PanSplat & GT Crop
            \end{tabular}
        }
        \vspace{-0.5em}
        \caption{
            \textbf{Qualitative comparisons on synthetic datasets.}
            We show the input panorama pairs and the ground truth novel views on the left, and compare the zoomed-in results on the right to highlight the differences.
            Our PanSplat generates overall sharper images with \textcolor{orange}{more high-frequency details} and \textcolor{cyan}{improved geometry}.
        }\label{fig:cmp_syn}
        \vspace{-1em}
    \end{figure*}
}

\newcommand{\row}[3]{
    \makecell{
        \includegraphics[width=0.14\linewidth]{results/#1/in_0/#2.jpg}
        \\
        \includegraphics[width=0.14\linewidth]{results/#1/in_1/#2.jpg}
    } &
    \makecell{
        \includegraphics[width=0.293\linewidth]{results/#1/gt_bbox/#2.jpg}
    } &
    \makecell{
        \includegraphics[width=0.1465\linewidth, clip, trim={#3}]{results/#1/panogrf/#2.jpg}
    } &
    \makecell{
        \includegraphics[width=0.1465\linewidth, clip, trim={#3}]{results/#1/mvsplat/#2.jpg}
    } &
    \makecell{
        \includegraphics[width=0.1465\linewidth, clip, trim={#3}]{results/#1/pansplat/#2.jpg}
    } &
    \makecell{
        \includegraphics[width=0.1465\linewidth, clip, trim={#3}]{results/#1/gt/#2.jpg}
    }
    \\
}

\newcommand{\dataset}[3]{
    \multirow{2}{*}{\raisebox{-9ex}{\rotatebox[origin=c]{90}{#1}}} & #2
    & #3
}

\cmponerow{
    \dataset{Matterport3D}{
        \row{cmp_mp3d}{0}{239px 190px 593px 130px}
    }{
        \row{cmp_mp3d}{7}{504px 115px 270px 147px} 
    }
    \dataset{Replica}{
        \row{cmp_replica}{8}{0px 20px 844px 312px} 
    }{
        \row{cmp_replica}{1}{609px 40px 203px 260px}
    }
    \dataset{Residential}{
        \row{cmp_residential}{1}{852px 150px 0px 190px}
    }{
        \row{cmp_residential}{2}{380px 120px 464px 212px}
    }
}

}

\subsection{Experimental Setup}

\noindent\textbf{Datasets.}
For comparison with existing methods, we evaluate PanSplat on three synthetic datasets: Matterport3D~\cite{chang2017matterport3d}, Replica~\cite{straub2019replica}, and Residential~\cite{habtegebrial2022somsi}, all at a resolution of $512 \times 1024$.
We follow the data split of PanoGRF~\cite{chen2023panogrf} to train on Matterport3D with a baseline (distances between input views) of 1.0, and evaluate using fixed baselines of 1.0, 1.5, and 2.0 meters.
For Replica and Residential, the baselines are 1.0 and approximately 0.3 meters, respectively.
In each case, a middle view is used as the target for both training and evaluation.
To scale up to 4K resolution on synthetic data, we render a 4K dataset using Matterport3D.
For real-world fine-tuning at 4K resolution, we utilize the 360Loc~\cite{huang2024360loc} dataset and a self-captured Insta360 dataset.
360Loc contains posed panorama sequences across four scenes, with an average baseline of 0.47 meters.
We select one scene as test set and fine-tune PanSplat on the other three scenes, using two views spaced two frames apart as input and evaluating across all four views.
We record two videos walking through indoor and outdoor scenes at 24 FPS using a 360° camera (Insta360 X4).
For camera pose estimation, we use OpenVSLAM~\cite{sumikura2019openvslam} without loop closure.
From this dataset, we select two views spaced 15 frames apart as input, evaluating all 17 frames.

\noindent\textbf{Implementation Details.}
We first train the model on Matterport3D at a height of 256, then fine-tune it at 512.
For 4K fine-tuning, we progressively increase the height from 1024 to 2048, with deferred backpropagation enabled.
For real datasets, we fine-tune on 360Loc, incrementally raising the resolution from 512 to 2048.

\noindent\textbf{Evaluation Metrics.}
Following PanoGRF~\cite{chen2023panogrf}, we use PSNR, SSIM~\cite{hore2010image}, LPIPS~\cite{zhang2018unreasonable}, and WS-PSNR~\cite{sun2017weighted} to evaluate image quality, but focus more on WS-PSNR as it considers pixel density of equirectangular images.

\subsection{Comparison with Previous Works}\label{sec:cmp}

\noindent\textbf{Baselines.}
We compare PanSplat with several feed-forward methods, including PanoGRF~\cite{chen2023panogrf}, NeuRay~\cite{liu2022neural}, IBRNet~\cite{wang2021ibrnet}, and OmniSyn~\cite{li2022omnisyn}, as well as with an optimization-based method, S-NeRF (PanoGRF's spherical adaption of NeRF~\cite{mildenhall2020nerf}), all at a resolution of $512 \times 1024$.
Notably, PanoGRF requires 23.8 seconds to generate an image, whereas PanSplat achieves the same result in just 0.34 seconds (0.32 seconds for the feed-forward network inference and 0.02 seconds for 3DGS rendering), making it up to $\bm{70 \times}$ faster. PanSplat enables real-time rendering and achieves a speed that is not feasible for NeRF-based approaches.
To compare with the latest 3DGS techniques, we adapt MVSplat~\cite{chen2025mvsplat}, a feed-forward method designed for perspective images, by replacing its camera model with a spherical camera and following their protocol to train on Matterport3D and fine-tune on 360Loc.
We follow the evaluation protocol of PanoGRF and report their original results of PanoGRF, NeuRay, IBRNet, OmniSyn and S-NeRF.

\noindent\textbf{Quantitative Results.}
\cref{tbl:cmp_syn} presents a quantitative comparison on Matterport3D, the dataset all methods are trained on.
PanSplat consistently outperforms all competing methods, not only on the training baseline of 1.0 meters but also when generalizing to wider baselines of 1.5 and 2.0 meters.
Although MVSplat serves as a strong baseline with recent advancements in 3DGS, it falls notably short of PanSplat's performance, especially at wider baselines.
To further evaluate generalization, we test on Replica and Residential datasets, where PanSplat achieves the best performance across most metrics, highlighting its strong generalization capability.
For real-world datasets where depth ground truth is unavailable, we compare with MVSplat, a method also supports training without depth supervision.
As shown in \cref{tbl:cmp_real}, PanSplat consistently outperforms MVSplat across all metrics, with deferred blending (w/ Deferred BL) providing an additional performance boost.

\noindent\textbf{Qualitative Results.}
\cref{fig:cmp_syn} presents qualitative comparisons on synthetic datasets, where we compare PanSplat with the best-performing baselines, PanoGRF and MVSplat.
Overall, Gaussian-based methods (PanSplat and MVSplat) preserve more details and produce sharper images compared to PanoGRF.
Furthermore, thanks to the spherical 3D Gaussian pyramid, PanSplat generates \textcolor{orange}{more detailed textures, particularly in high-frequency areas} such as the pattern on the wall in the first and second rows, and the blinds in the fifth row.
Besides, the use of a hierarchical spherical cost volume enables \textcolor{cyan}{more accurate depth estimation, resulting in improved geometry} in other samples.

{
\newcommand{\cmponerow}[1]{
    \begin{figure*}[t]
        \centering
        \small
        \setlength\tabcolsep{1px}
        \resizebox{1.\linewidth}{!}{
            \begin{tabular}{cccccc}
                #1
                Input Views & GT View & Base & +Fibo & +3DGP (Full) & GT Crop
            \end{tabular}
        }
        \vspace{-0.5em}
        \caption{
            \textbf{Qualitative comparisons of ablation study.}
            Our Fibonacci Gaussians (+Fibo) reduces Gaussian count without compromising image quality, and our 3D Gaussian Pyramid (+3DGP) further enhances quality.
        }\label{fig:ablation}
        \vspace{-1.2em}
    \end{figure*}
}

\newcommand{\row}[2]{
    \makecell{
        \includegraphics[width=0.14\linewidth]{results/cmp_mp3d/in_0/#1.jpg}
        \\
        \includegraphics[width=0.14\linewidth]{results/cmp_mp3d/in_1/#1.jpg}
    } &
    \makecell{
        \includegraphics[width=0.293\linewidth]{results/ab_mp3d/gt_bbox/#1.jpg}
    } &
    \makecell{
        \includegraphics[width=0.1465\linewidth, clip, trim={#2}]{results/ab_mp3d/wo_pgs_fibo/#1.jpg}
    } &
    \makecell{
        \includegraphics[width=0.1465\linewidth, clip, trim={#2}]{results/ab_mp3d/wo_pgs/#1.jpg}
    } &
    \makecell{
        \includegraphics[width=0.1465\linewidth, clip, trim={#2}]{results/cmp_mp3d/pansplat/#1.jpg}
    } &
    \makecell{
        \includegraphics[width=0.1465\linewidth, clip, trim={#2}]{results/cmp_mp3d/gt/#1.jpg}
    }
    \\
}

\cmponerow{
    \row{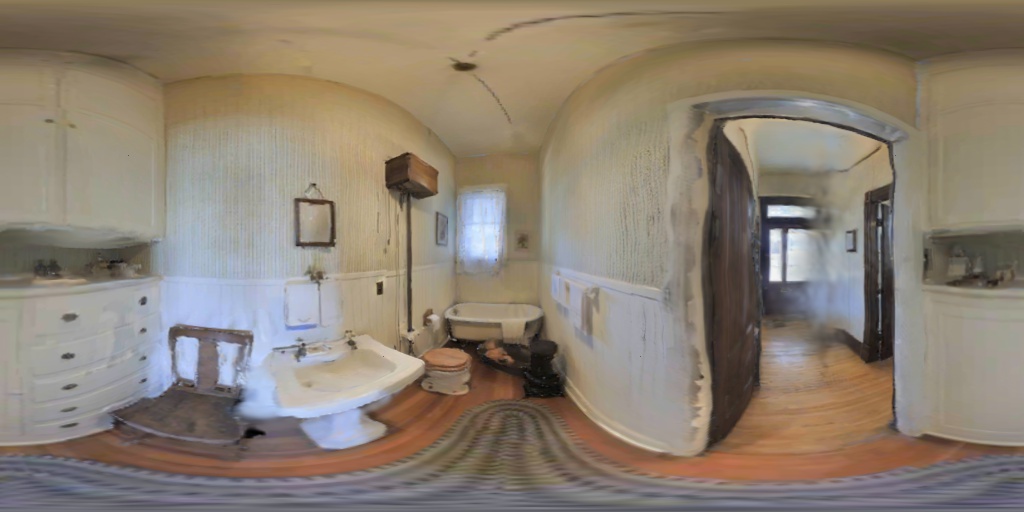}{559px 150px 273px 170px}
    \row{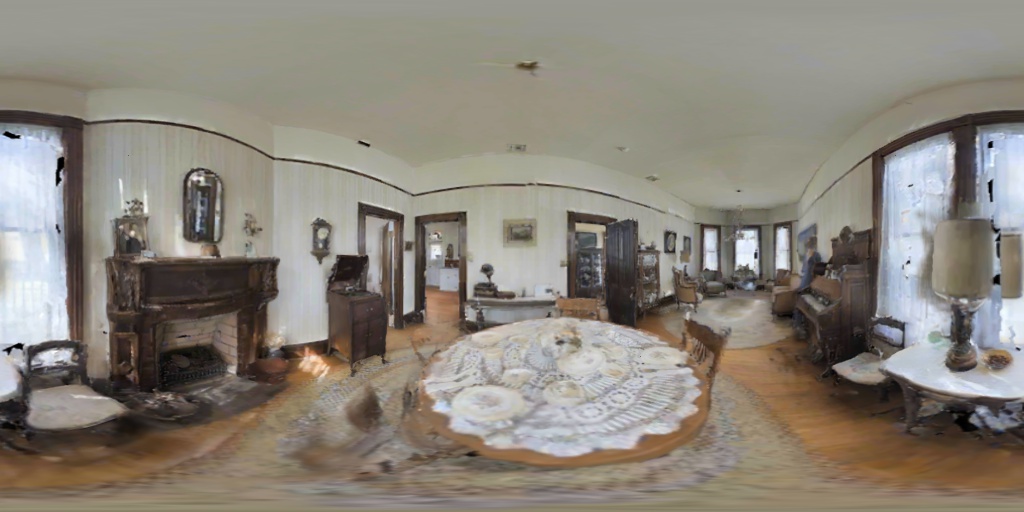}{449px 40px 363px 260px}
}

}

\subsection{Ablation Study}\label{sec:ablation}

\noindent\textbf{Synthetic Datasets.}
We conduct an ablation study on Matterport3D to assess the impact of the two key components: Fibonacci Gaussians and 3D Gaussian pyramid.
As shown in \cref{tbl:ablation}, we begin by evaluating a baseline model (Base) without these components, utilizing a single 1/4-resolution cost volume to maintain comparable computational cost and memory usage with the full model.
Next, we add Fibonacci Gaussians (+Fibo), which significantly reduces the number of Gaussians without compromising image quality.
Finally, we incorporate 3D Gaussian Pyramid (+3DGP) to capture multi-scale details, resulting in further performance improvements.
Although 3DGP introduces additional Gaussians, the use of Fibo helps offset the increase, leading to an overall reduction in the total Gaussian count.
\cref{fig:ablation} presents visual comparisons, where the baseline model fails to capture fine details, whereas the full model with 3DGP generates sharper images with more accurate geometry.
It also demonstrates that the use of Fibo does not introduce visible artifacts, highlighting its effectiveness in reducing the number of Gaussians without sacrificing quality.

\begin{table}[t]
    \setlength\tabcolsep{2pt}
    \centering
    \begin{tabular}{ccccc}
        \toprule
        Setup & \#Gaussian (K) & WS-PSNR$\uparrow$ & SSIM$\uparrow$ & LPIPS $\downarrow$ \\
        \cmidrule(r){1-1}\cmidrule(r){2-2}\cmidrule(r){3-5}
        Base & 1,049 (100\%) & 27.07 & 0.895 & 0.127 \\ 
        +Fibo & 668 (63.67\%) & \cellcolor{orange!30}27.86 & \cellcolor{orange!30}0.906 & \cellcolor{orange!30}0.116 \\ 
        +3DGP (Full) & 887 (84.55\%) & \cellcolor{red!30}28.81 & \cellcolor{red!30}0.931 & \cellcolor{red!30}0.091 \\ 
        \bottomrule
    \end{tabular}
    \vspace{-0.5em}
    \caption{
        \textbf{Ablation study.}
        We count the number of Gaussians (\#Gaussians) and evaluate performance on the Matterport3D 1.0m baseline test set.
        We progressively add our two proposed components to the base model and measure the performance.
    }\label{tbl:ablation}
    \vspace{-1em}
\end{table}

\noindent\textbf{Real Datasets.}
In \cref{sec:cmp} and \cref{tbl:cmp_real}, we demonstrate that deferred blending (w/ Deferred BL) substantially enhances the performance on real-world datasets.
A more detailed analysis of deferred blending’s impact is provided in \cref*{sec:ext_to_real} of the supplementary material.

\begin{figure}[t]
    \centering
    \includegraphics[width=1.0\linewidth, trim={0 0 0 0}, clip]{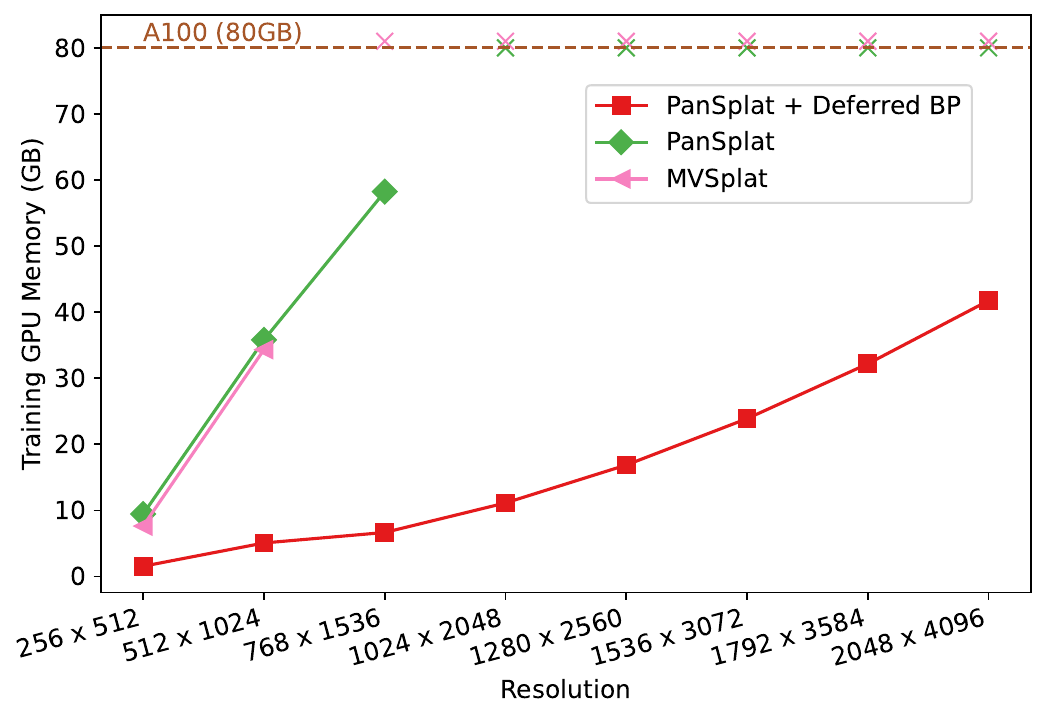}
    \vspace{-2em}
    \caption{
        \textbf{Training GPU memory consumption at different resolutions}, 
        where $\times$ indicates out-of-memory errors even on a 80GB A100.
        Memory consumption is tested with a batch size of 1.
    }\label{fig:gpu_memory}
    \vspace{-1em}
\end{figure}

\noindent\textbf{Scaling Up to 4K Resolution.}
We evaluate the impact of the two-step deferred backpropagation on training memory consumption in \cref{fig:gpu_memory}.
As shown, MVSplat reaches memory overflow at a relatively low resolution of $512 \times 1024$, while PanSplat is able to support $768 \times 1536$ resolutions due to its fixed cost volume size and the efficient design of the Gaussian heads.
It is worth noting that this design choice does not compromise image quality, as discussed in \cref{sec:cmp}; rather, it enables deferred backpropagation, drastically reducing memory consumption during training and allowing PanSplat to support 4K resolution on a single A100 GPU.
We present qualitative 4K results of PanSplat in~\cref{fig:teaser} and include additional results in the supplementary video.
We also provide an in-depth analysis of design choices and inference memory usage in~\cref*{sec:4k} of the supplementary material, which shows that PanSplat can inference at 4K resolution on a 24GB RTX 3090 GPU.

\section{Conclusion}\label{sec:conclusion}
In this paper, we have presented PanSplat, a novel generalizable, feed-forward approach for novel view synthesis from wide-baseline panoramas.
To efficiently support 4K resolution ($2048 \times 4096$) for immersive VR applications, we have introduced a pipeline that enables two-step deferred backpropagation.
In addition, we have proposed a spherical 3D Gaussian pyramid with a Fibonacci lattice arrangement tailored for panorama formats, to enhance both rendering quality and efficiency.
Extensive experiments have demonstrated the superiority of PanSplat over existing techniques in terms of image quality and resolution.

\noindent\textbf{Limitations.}
While PanSplat provides a promising solution for high-resolution panoramic novel view synthesis, it lacks support for dynamic scenes with moving objects, a frequent requirement in real-world applications.
Future work could explore extending PanSplat to handle dynamic scenes by incorporating motion-aware representations.

\noindent\textbf{Acknowledgement:}
This research is supported by Building 4.0 CRC.

{
    \small
    \bibliographystyle{ieeenat_fullname}
    \bibliography{main}
}

\clearpage
\setcounter{page}{1}
\maketitlesupplementary

\counterwithin{figure}{section}
\counterwithin{table}{section}
\renewcommand\thesection{\Alph{section}}
\renewcommand\thetable{\thesection.\arabic{table}}
\renewcommand\thefigure{\thesection.\arabic{figure}}
\setcounter{section}{0}

The supplementary material is organized as follows.
In~\cref{sec:arch}, we provide additional details on the network architectures.
In \cref{sec:defbp}, we provide additional details on the Gaussian parameter prediction and rendering.
In~\cref{sec:exp_details}, we provide additional details on the experiment settings.
In~\cref{sec:narrow_baseline}, we provide quantitative comparisons on narrow baselines.
In~\cref{sec:more_ablation}, we provide more ablation studies.
In~\cref{sec:ext_to_real}, we provide details on extending to real data.
In~\cref{sec:4k}, we provide details on scaling up to 4K resolution.
Finally, in~\cref{sec:demo}, we provide details on the demo video.

\section{Network Architectures}\label{sec:arch}
In \cref*{sec:method} of the main paper, we present our PanSplat architecture in two parts: the Hierarchical Spherical Cost Volume (\cref*{sec:cost_volume}) and the Gaussian Heads (\cref*{sec:gaussian_pred}).
Here, we provide additional details on the network architectures.

\noindent\textbf{Hierarchical Spherical Cost Volume.}
For feature pyramid extraction, we adopt a FPN architecture~\cite{lin2017feature} enhanced with a Swin Transformer~\cite{liu2021swin}.
The Swin Transformer consists of 6 Transformer blocks, each with a self-attention layer and a cross-view attention layer.
We use the xFormers~\cite{xFormers2022} library for the transformer-based network for better efficiency.
We apply Swin Transformer to the coarsest level of the feature map from the FPN encoder, then upsample the feature map to different levels with the FPN decoder.
The result is a feature pyramid with 4 levels, with channel dimensions of 128, 96, 64, 32 from the coarsest to the finest level.
For hierarchical spherical cost volume refinement, we adopt a 2DU-Net~\cite{chen2025mvsplat} with cross-view attention at the bottleneck layer for each level.
We set depth candidates to 128, 64, 32 and channel dimensions of 2D U-Net to 128, 64, 32 for each level, respectively.

\noindent\textbf{Gaussian Heads.}
We adopt a lightweight 3-layer CNN architecture for each Gaussian head, with a kernel size of $3 \times 3$ and a stride of 1, to extract feature map $\bm{\tilde{F}}^{l}_{i}$ for each view ${i}$ at level ${l}$.
We then sample a feature vector from the feature maps for each Gaussian, based on the pixel location defined on the Fibonacci lattice.
Finally, a linear layer is applied to predict the Gaussian parameters $( \bm{\mu}^{l}_{i}, \bm{\alpha}^{l}_{i}, \bm{\Sigma}^{l}_{i}, \bm{c}^{l}_{i} )$.
Specifically, to estimate Gaussian centers $\bm{\mu}^{l}_{i}$, we first estimate the correlation vectors $\bm{c}^{l}_{i}$, then apply the same operations used for the cost volume to get a depth, which is then unprojected to 3D coordinates as mentioned in \cref*{sec:s3dgp} of the main paper.
The opacity $\bm{\alpha}^{l}_{i}$ is predicted as a scalar value, followed by a $\text{sigmoid}$ activation to normalize it to $[0, 1]$.
The covariance $\bm{\Sigma}$ is composed of scaling vectors and quaternions, where the scaling is calculated as predicted normalized vectors $\bm{s}^{l}_{i} \in [{s}_{\text{min}}, {s}_{\text{max}}]$ multiplied by the pixel size.
This restricts the Gaussian to a similar scale as the pixel, accounting for the change in pixel size across different levels.
The color $\bm{c}^{l}_{i}$ is represented as spherical harmonic coefficients.

\section{Gaussian Parameter Prediction and Rendering Details}\label{sec:defbp}

\begin{figure*}[t]
    \vspace{-1em}
    \centering
    \includegraphics[width=1.\linewidth, trim={0 0 0 0}, clip]{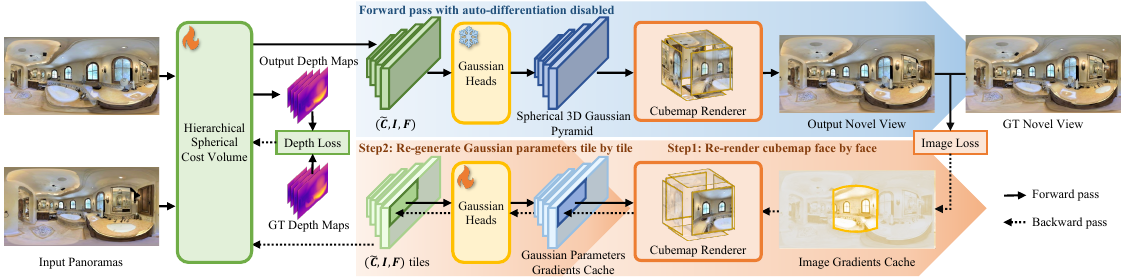}
    \vspace{-1.5em}
    \caption{
        \textbf{Two-step deferred backpropagation.}
        We propose a training strategy tailored for high-resolution panorama novel view synthesis.
        See~\cref{sec:defbp} for details.
        \emph{For simplicity, intermediate results of only a single view are shown.}
    }\label{fig:defbp}
    \vspace{-1em}
\end{figure*}

In \cref*{sec:gaussian_pred} of the main paper, we introduce Gaussian heads with local operations and a cubemap renderer.
Based on these two components, we propose a two-step deferred backpropagation technique to enable training at 4K resolution.
Here, we provide additional details on the deferred backpropagation technique, as shown in~\cref{fig:defbp}, as well as the two components it relies on.

\noindent\textbf{Tiled Operation for Gaussian Heads.}
We mentioned in~\cref*{sec:gaussian_pred} of the main paper that we exploit the local operations in the Gaussian heads to enable tiled operation for inference and deferred backpropagation.
To be more specific, the inputs to the Gaussian heads on different levels are evenly split into ${N} \times {N}$ tiles, then fed into the Gaussian heads separately.
However, this naive tiled operation impacts the boundary value of the output tiles, due to the zero padding of each convolutional layer, leading to discontinuity at the tile boundaries.
Instead, we refine this design to output results identical to the non-tiled operation with a pre-padding operation.
First the inputs are padded by 3 pixels, to accommodate the field of perception of the Gaussian heads.
The padding involves copying the border pixels of left and right sides to the opposite side, which ensures loop continuity of the spherical geometry.
The top and bottom sides are padded with zeros.
Then, the tile regions are enlarged by 3 pixels to include the above padding, and introduce a 3-pixel overlap between adjacent tiles.
The output tiles are finally cropped to the original size, stitched to a continuous, full resolution output.

\noindent\textbf{Details of Cubemap Renderer.}
One key component of two-step deferred backpropagation is the cubemap renderer, which provides a differentiable rendering pipeline for the spherical 3D Gaussian pyramid.
As shown in~\cref{fig:defbp}, the cubemap renderer renders 6 faces (front, back, left, right, top, bottom) of the cubemap separately, then stitches them into an equirectangular panorama.
This allows sequential face rendering for memory efficiency or batched face rendering for speedup.
We build the cubemap renderer based on the CUDA 3DGS renderer~\cite{kerbl20233d} that implements with perspective camera projection.
After rendering each face, we apply a bilinear grid sampling to stitch the faces into an equirectangular panorama.
Specifically, the coordinates of pixels in the equirectangular panorama are first transformed to the corresponding coordinates on the cubemap image.
Then the pixel values are sampled from the cubemap image using bilinear interpolation.
To achieve seamless stitching, we pad the edge pixels of the adjacent 4 faces to each face, ensuring the pixels interpolated on the edge have correct neighboring pixels from two nearby faces.

\noindent\textbf{Details of Two-step Deferred Backpropagation.}
As shown in~\cref{fig:defbp}, the two-step deferred backpropagation consists of a forward pass and two deferred backpropagation steps.
Before the forward pass, we construct the hierarchical spherical cost volume with auto-differentiation on, and preserves the computational graph throughout the training step for efficiency.
Then we disable auto-differentiation for a forward pass to render the full panorama.
The full panorama is used for computing an image loss, with auto-differentiation on, to backpropagate and cache gradients to the image.
Subsequently, we enable auto-differentiation and backpropagate gradients in two steps.
In step one, the panorama is re-rendered face by face as cubemap to backpropagate and accumulate gradients to the Gaussian parameters.
In step two, the Gaussian parameters are re-generated tile by tile, with gradients backpropagated and accumulated to the network parameters.
Additionally, the gradients from the depth loss are accumulated to the network together with the gradients from the image loss.
When training on real datasets without ground truth depth, the depth loss is replaced by auxiliary Gaussian heads and image loss as discussed in~\cref*{sec:training} of the main paper.
In~\cref{sec:4k}, we provide more details on how the two-step deferred backpropagation saves memory consumption during training.

\section{Experiment Details}\label{sec:exp_details}

\noindent\textbf{High-resolution Synthetic Datasets.}
For synthetic data, we use the low-resolution ($512 \times 1024$) synthetic datasets Matterport3D~\cite{chang2017matterport3d}, Replica~\cite{straub2019replica}, and Residential~\cite{habtegebrial2022somsi} rendered by PanoGRF~\cite{chen2023panogrf}.
Additionally, we render two high-resolution datasets ($1024 \times 2048$ / $2048 \times 4096$) using Matterport3D for fine-tuning.
Specifically, we follow PanoGRF's rendering protocol to render 6 perspective images at $512 \times 512$ / $1024 \times 1024$ resolution respectively on the cubemap faces, then stitch them into an equirectangular panorama image.
We render 2 views with a baseline of 1.0 meter as input, and 1 view in the middle as the target view.
The two datasets contain 5,000 / 2,000 samples for training.
We render the test set in consistent with PanoGRF, with 10 samples for each dataset, which are used for demonstration in the demo video.

\noindent\textbf{High-resolution Real Datasets.}
We use two real-world datasets to demonstrate generalization to real-world scenarios.
For fine-tuning to real images, we use the 360Loc~\cite{huang2024360loc} dataset as it provides accurate pose registration from dense point cloud reconstructions and lidar scans.
In addition, it is the largest dataset with high-resolution panoramic image sequences as far as we know, with 18 sequences (12 daytime and 6 nighttime) across 4 scenes, totaling 9,334 frames.
We select one scene with 5 sequences as the test set, and fine-tune on the other 3 scenes with 13 sequences.
When fine-tuning, we randomly sample two views with varying baselines spaced 1 to 4 frames apart and select a target view in between. 
During evaluation, we select two views spaced 2 frames apart as input, and use all 4 views as the target to calculate the metrics.
For analyzing image quality over different frame distances in \cref{sec:ext_to_real}, we find that 360Loc is too sparse (average baseline of 0.47 meters) to provide a reasonable amount of frame distance samples.
Therefore, we also capture a high-resolution Insta360 dataset with two sequences (one indoor and one outdoor) totaling 38K frames.
Insta360 is recorded at 8K resolution and 24 FPS, later down-sampled to 4K for evaluation.
We use OpenVSLAM~\cite{sumikura2019openvslam} for camera pose estimation, disabling loop closure to avoid bad loop detection in repetitive environments.
For evaluation purposes, we select two views spaced 15 frames apart as input, and evaluate all 17 frames.
For evaluation on both datasets, we evenly sample 100 pairs of input views for each sequence, and average the results over all target views.

\stepcounter{section}
\begin{table*}[t]
    \setlength\tabcolsep{4pt}
    \centering
    \begin{tabular}{ccccccccc}
        \toprule
        Baseline & \multicolumn{4}{c}{0.2m} & \multicolumn{4}{c}{0.5m} \\
        \cmidrule(lr){1-1}\cmidrule(lr){2-5}\cmidrule(lr){6-9}
        Method &PSNR$\uparrow$ & WS-PSNR$\uparrow$ & SSIM$\uparrow$ & LPIPS$\downarrow$ &PSNR$\uparrow$&  WS-PSNR$\uparrow$ & SSIM$\uparrow$ & LPIPS$\downarrow$ \\
        \cmidrule(lr){1-1}\cmidrule(lr){2-5}\cmidrule(lr){6-9}
        S-NeRF  & 20.79 & 19.52 & 0.697 & 0.376 & 17.95 & 16.81 & 0.628 & 0.486 \\
        OmniSyn  & 28.95 & 28.26 & 0.913 & 0.180 & 26.59 & 26.07 & 0.890 & 0.201 \\
        IBRNet  & 30.53 & 29.63 & 0.927 & 0.136 & 28.22 & 27.26 & 0.884 & 0.199\\
        NeuRay  & \cellcolor{yellow!30}33.54 & 32.33 & 0.949 & 0.107 & 30.88 & 29.81 & 0.920 & 0.154 \\
        PanoGRF & \cellcolor{red!30}34.29 & \cellcolor{red!30}33.27 & \cellcolor{yellow!30}0.952 & \cellcolor{yellow!30}0.098 & \cellcolor{yellow!30}31.41 & \cellcolor{yellow!30}30.46 & \cellcolor{yellow!30}0.924 & \cellcolor{yellow!30}0.132  \\
        MVSplat & 32.93 & \cellcolor{yellow!30}32.04 & \cellcolor{orange!30}0.955 & \cellcolor{red!30}0.063 & \cellcolor{orange!30}31.55 & \cellcolor{orange!30}30.58 & \cellcolor{orange!30}0.943 & \cellcolor{orange!30}0.075 \\ 
        PanSplat & \cellcolor{orange!30}33.92 & \cellcolor{orange!30}32.88 & \cellcolor{red!30}0.959 & \cellcolor{orange!30}0.066 & \cellcolor{red!30}32.46 & \cellcolor{red!30}31.42 & \cellcolor{red!30}0.950 & \cellcolor{red!30}0.072 \\ 
        \bottomrule
    \end{tabular}
    \vspace{-0.5em}
    \caption{
        \textbf{Quantitative comparison on narrow baselines.}
        We compare on Matterport3D under the baseline of 0.2 and 0.5 meters.
        Top results are highlighted in \colorbox{red!30}{top1}, \colorbox{orange!30}{top2}, and \colorbox{yellow!30}{top3}.
    }\label{tbl:cmp_narrow}
\end{table*}
\addtocounter{section}{-1}

\noindent\textbf{Implementation Details.}
We set the number of depth candidates ${D}$ for the coarsest level to 128.
Our model is implemented in PyTorch and trained on a single 80GB NVIDIA A100 GPU using the Adam optimizer with a learning rate of $2 \times 10^{-4}$.
We use the pre-trained weights of UniMatch~\cite{xu2023unifying} to initialize the Swin Transformer of feature pyramid extractor.
We also load the pre-trained weights of the monocular depth model~\cite{jiang2021unifuse} trained by PanoGRF~\cite{chen2023panogrf} and freeze it during training.
Initially, we train the model on Matterport3D with an image height of 256 and a batch size of 6 for 10 epochs, then fine-tune it with an image height of 512 and a batch size of 2 for 5 epochs.
For 4K Matterport3D fine-tuning, we gradually increase the resolution from a height of 1024 to 2048 over 3 epochs at each stage.
To fine-tune on 4K 360Loc, we incrementally raise the resolution from a height of 512 to 1024 and finally 2048, with 65K, 26K, and 13K iterations for each stage, respectively.
At resolutions of 1024 and 2048, we enable two-step deferred backpropagation with 4 and 16 tiles, setting batch sizes to 3 and 1, respectively.
When fine-tuning on 360Loc at 1024 and 2048, we freeze the hierarchical spherical cost volume and only fine-tune the Gaussian heads.
During evaluation, we generalize the model directly from Matterport3D to the Replica and Residential, and from 360Loc to the Insta360 dataset, without additional fine-tuning.

\section{Quantitative Comparisons on Narrow Baselines}\label{sec:narrow_baseline}

We follow the evaluation protocol of PanoGRF~\cite{chen2023panogrf} to further evaluate on generalization to narrow baselines on Matterport3D.
As shown in \cref{tbl:cmp_narrow}, while PanSplat achieves the best performance at the 0.5m baseline, it also demonstrates competitive results at the 0.2m baseline, indicating strong generalization across varying baseline distances.

\section{More Ablation Studies}\label{sec:more_ablation}

\begin{table}[t]
    \setlength\tabcolsep{2pt}
    \centering
    \begin{tabular}{cccc}
        \toprule

        Setup &  WS-PSNR$\uparrow$ & SSIM$\uparrow$ & LPIPS $\downarrow$ \\
        \cmidrule(r){1-1}\cmidrule(r){2-4}
        w/o Mono depth & 28.84 & 0.929 & 0.092 \\ 
        w/o 3DGP residual & 28.14 & 0.922 & 0.102 \\ 
        w/o Hierarchical CV & 26.95 & 0.857 & 0.180 \\ 
        w/o First three GHs & 28.05 & 0.919 & 0.105 \\ 
        Full & 28.81 & 0.931 & 0.091 \\ 
        \bottomrule
    \end{tabular}
    \caption{
        \textbf{Full ablation study.}
        We evaluate the impact of certain design choices on PanSplat's performance.
        Mono depth refers to integrating monocular depth feature from PanoGRF~\cite{chen2023panogrf} to the hierarchical spherical cost volume, which is not our contribution and is insignificant to performance, but we include it in the Full model for the best results.
        Other design choices are ablated from the Full model, and significantly affect the performance.
    }\label{tbl:ablation_all}
\end{table}

{

\newcommand{\row}[2]{
    \raisebox{8.5ex}{\rotatebox[origin=c]{90}{#1}} &
    \includegraphics[width=0.24\linewidth, clip]{results/pgs/#2/stage_0.jpg} &
    \includegraphics[width=0.24\linewidth, clip]{results/pgs/#2/stage_1.jpg} &
    \includegraphics[width=0.24\linewidth, clip]{results/pgs/#2/stage_2.jpg} &
    \makebox[0.24\linewidth]{
        \IfFileExists{results/pgs/#2/stage_3.jpg}
        {
            \includegraphics[width=0.24\linewidth, clip]{results/pgs/#2/stage_3.jpg}
        }
        {
            \raisebox{8.5ex}{No stage \#3}
        }
    }
    \\
}

\begin{figure*}[t]
    \centering
    \scriptsize
    \setlength\tabcolsep{1px}
    \begin{tabular}{rcccc}
        \row{w/o 3DGP residual}{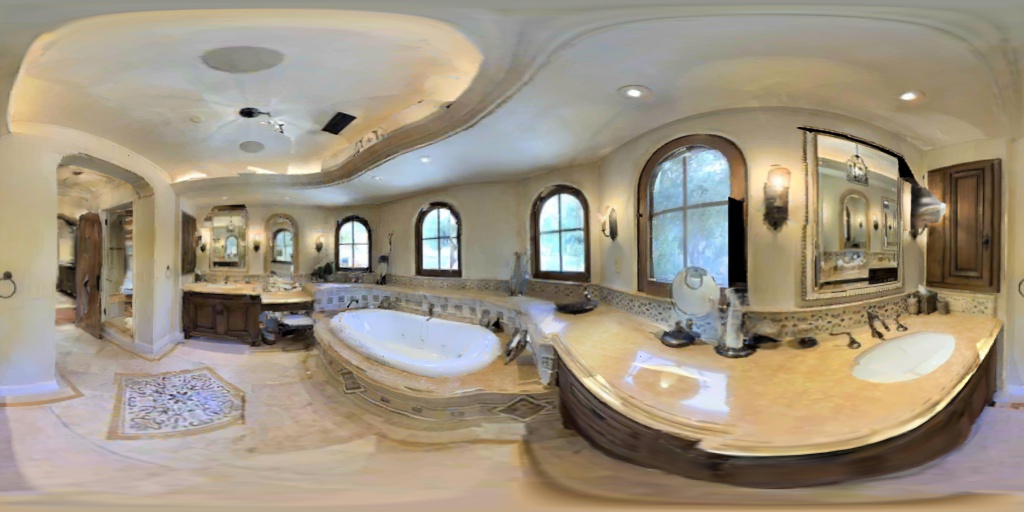}
        \row{w/o Hierarchical CV}{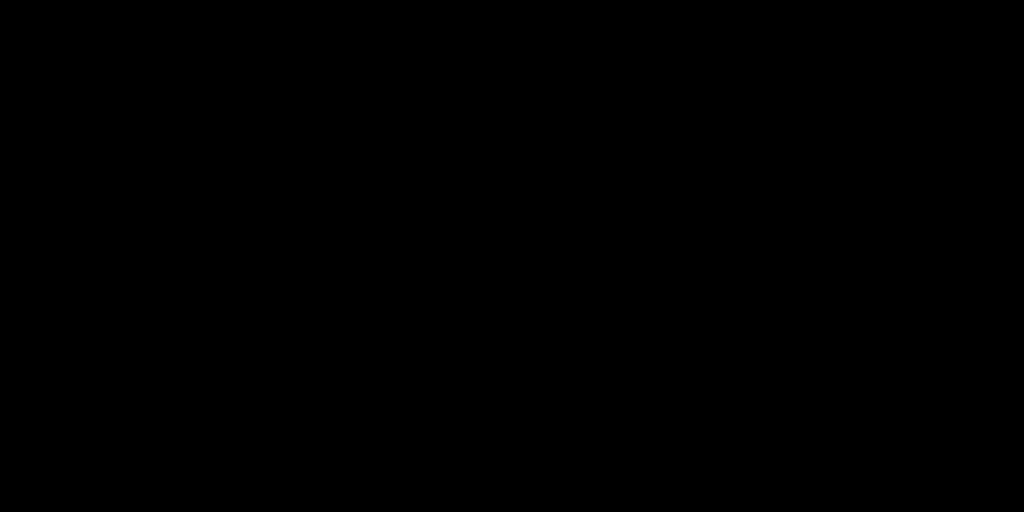}
        \row{Full}{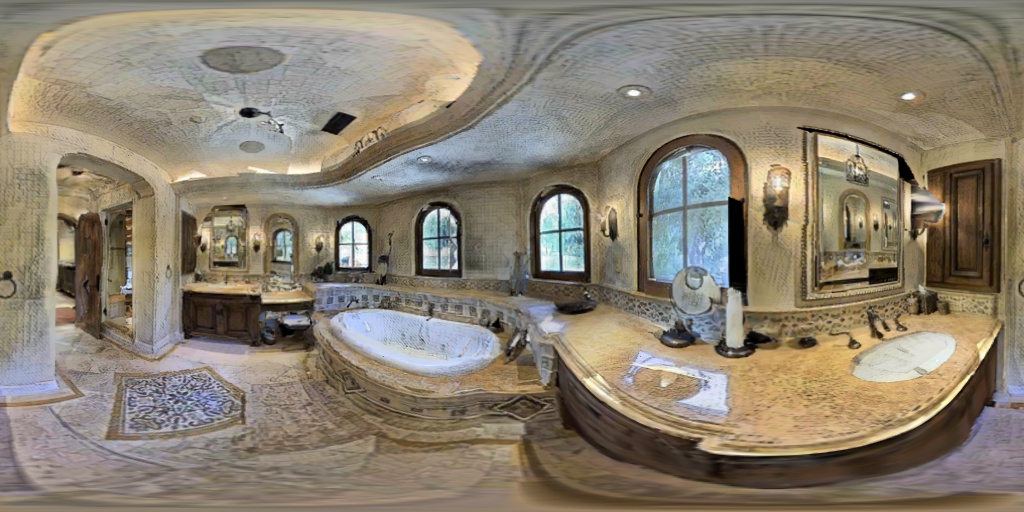}
        & Level \#3 & Level \#2 & Level \#1 & Level \#0
    \end{tabular}
    \vspace{-0.5em}
    \caption{
        \textbf{Visualization of 3D Gaussian Pyramid.}
        We visualize the rendering results of Gaussians from different levels of our 3D Gaussian Pyramid.
        Our full model (Full) successfully exploits the hierarchical structure of the 3D Gaussian Pyramid, where coarser levels mainly capture global structures and finer levels capture high-frequency details.
        In contrast, the ablated models (w/o 3DGP residual and w/o Hierarchical CV) fail to utilize all levels.
    }\label{fig:ablation_pgs}
    \vspace{-1em}
\end{figure*}

}

In \cref*{sec:ablation} of the main paper, we conduct an ablation study to analyze the contributions of Fibonacci Gaussians and the 3D Gaussian pyramid.
Here, we provide additional ablation studies in~\cref{tbl:ablation_all} and \cref{fig:ablation_pgs} to further analyze the impact of specific design choices in PanSplat.

\noindent\textbf{Monocular Depth Features.}
We first ablate the use of monocular depth features in the hierarchical spherical cost volume (w/o Mono depth) in~\cref{tbl:ablation_all}.
We note that integrating monocular depth features is a common practice in multi-view stereo methods~\cite{chen2023panogrf,xu2024depthsplat}.
Although in our case, the improvement is marginal, we include it in our final model for the best performance.

\noindent\textbf{Residual Design of 3D Gaussian Pyramid.}
Second, we ablate the residual design of the Gaussian heads (w/o 3DGP residual), which leads to a significant drop in performance.
To justify the performance gain from the residual design, we separately render the Gaussians from each level in~\cref{fig:ablation_pgs}.
It is shown that without the residual design, the coarsest two levels (Level \#3 and \#2) fail to output meaningful Gaussians, while the full model successfully distributes low frequency details to the coarser levels.
This demonstrates the effectiveness of the residual design in guiding the Gaussian heads to capture multi-scale details.

\stepcounter{section}
\begin{figure*}[t]
    \centering
    \includegraphics[width=1.0\linewidth, trim={0 0 0 0}, clip]{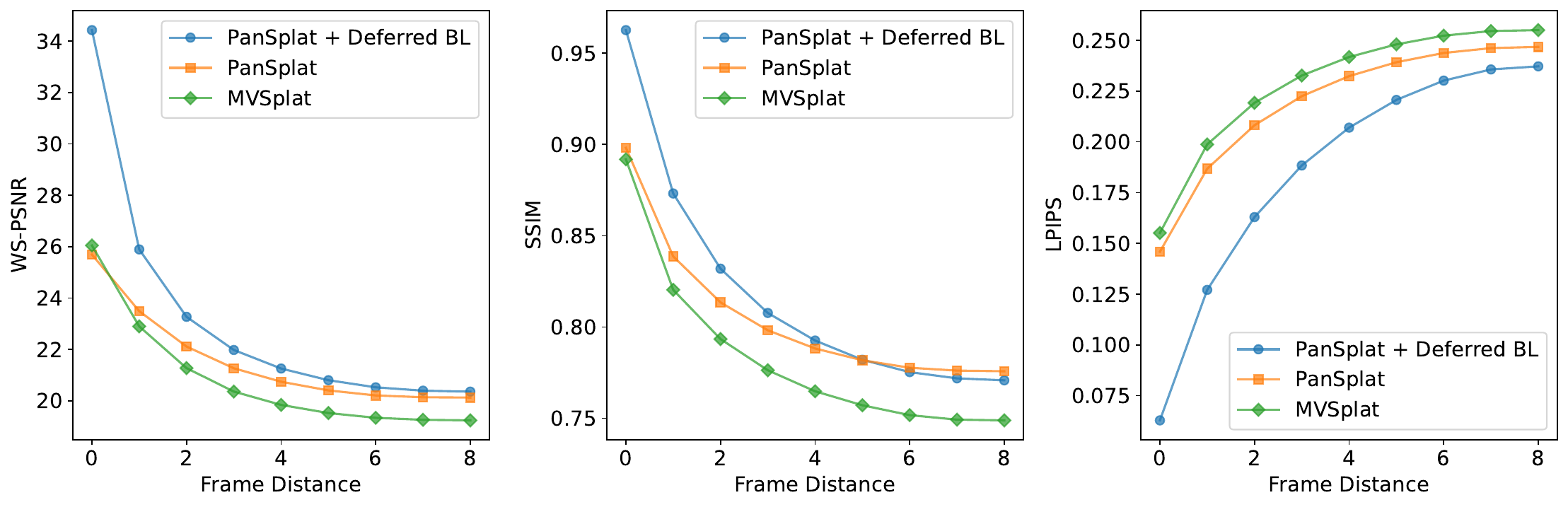}
    \vspace{-2em}
    \caption{
        \textbf{Quantitative comparisons on different frame distances.}
        We evaluate image quality metrics on Insta360 dataset with varying frame distances, comparing PanSplat with (PanSplat + Deferred BL) and without (PanSplat) deferred blending against MVSplat.
    }\label{fig:frame_vs_metric}
\end{figure*}

\addtocounter{section}{-1}

\noindent\textbf{Hierarchical Designs.}
Finally, we ablate the Hierarchical Cost Volume (w/o Hierarchical CV) and the First three Gaussian heads (w/o First 3 GH) respectively to analyze the joint impact of the two hierarchical designs.
Similar to~\cref*{sec:ablation} of the main paper, for w/o Hierarchical CV, we replace the hierarchical cost volume with a single 1/4-resolution cost volume with 128 depth candidates to maintain comparable computational cost and memory usage.
The removal of each of the two components hurts the performance significantly, indicating that the two hierarchical designs complement each other to achieve the best performance.
We find that w/o Hierarchical CV tends to fall into local minima where only the coarsest level is utilized, as shown in~\cref{fig:ablation_pgs}.

\section{Extending to Real Data}\label{sec:ext_to_real}

\noindent\textbf{Deferred Blending.}
In \cref*{sec:gaussian_pred} of the main paper, we introduce a deferred blending technique to mitigate artifacts from misaligned Gaussians due to moving objects and depth inconsistencies.
Here we provide additional details.
Specifically, on real datasets, instead of directly consolidating the Gaussians from two input views for rendering, we first separately render them from the same target view into two different images, which we denote as $\{\bm{\tilde{I}}_{i}\}_{{i} = 0}^{1}$.
Then we blend them based on the distances ${d}_{i}$ to the input views ${i}$ by:
\begin{equation}
    \bm{I} = \frac{{d}_{1} \bm{\tilde{I}}_{0} + {d}_{0} \bm{\tilde{I}}_{1}}{{d}_{0} + {d}_{1}}.
\end{equation}
The deferred blending aims to mitigate the influence of farther input view when rendering close to one of the input views, and relief the burden of matching moving objects.

\noindent\textbf{Experiments.}
To evaluate the impact of deferred blending, we analyze the relationship between image quality (WS-PSNR, SSIM, and LPIPS) and frame distance (the number of frames between the target view and the nearest input view) on the Insta360 dataset.
We compare PanSplat with (PanSplat + Deferred BL) and without (PanSplat) deferred blending, using MVSplat as a baseline.
As shown in~\cref{fig:frame_vs_metric}, PanSplat consistently outperforms MVSplat across all metrics and frame distances.
In addition, deferred blending provides notable performance gains, especially when the frame distance is small.
We further show visual comparisons on the 360Loc dataset in~\crefrange{fig:cmp_360loc_1}{fig:cmp_360loc_4} and on the Insta360 dataset in~\crefrange{fig:cmp_insta360_1}{fig:cmp_insta360_4}.
These results demonstrate that deferred blending significantly reduces artifacts arising from misaligned Gaussians (e.g., the dot pattern on the ceiling in~\cref{fig:cmp_insta360_2}) and moving objects (e.g., the camera operator at the bottom in~\cref{fig:cmp_360loc_2}).
It also provides nearly perfect results when rendering at the same location as one of the input views by isolating the influence of the farther input view.
This is particularly important for smooth transitions in virtual tours applications as shown in the demo video.

\section{Scaling Up to 4K Resolution}\label{sec:4k}

\begin{figure}[t]
    \centering
    \includegraphics[width=1.0\linewidth, trim={0 0 0 0}, clip]{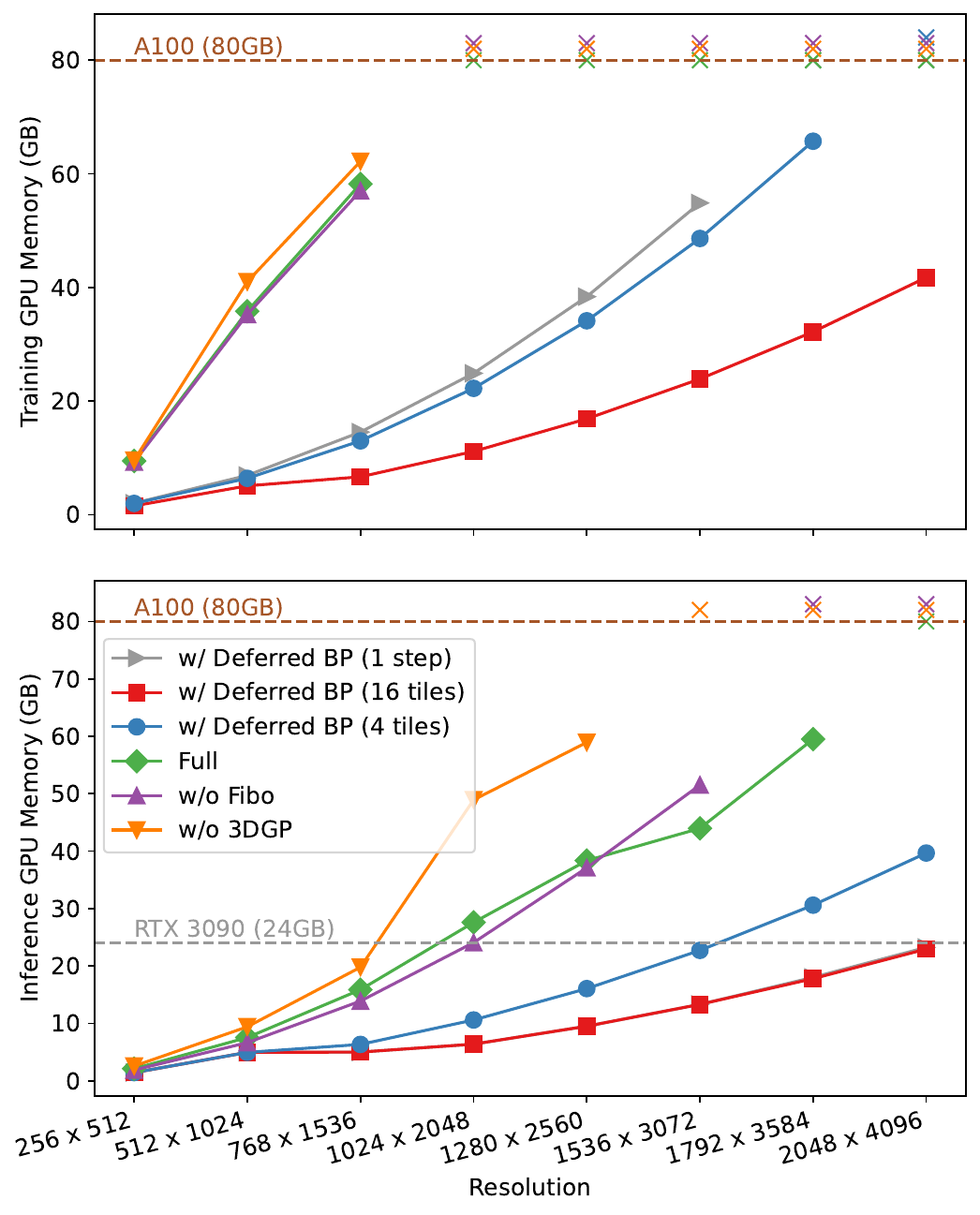}
    \vspace{-1.5em}
    \caption{
        \textbf{Full GPU memory consumption at different resolutions},
        where $\times$ indicates out-of-memory errors even on a 80GB A100.
        Note that w/ Deferred BP (1 step) is overlapped with w/ Deferred BP (16 tiles) for inference.
        Memory consumption is tested with a batch size of 1.
    }\label{fig:gpu_memory_ab}
\end{figure}

\addtocounter{section}{-1}
{

\newcommand{\row}[4]{
    \raisebox{14ex}{\rotatebox[origin=c]{90}{#1}} &
    \includegraphics[width=0.24\linewidth, clip, trim={#4}]{results/cmp_360loc/#2/#3/0.jpg} &
    \includegraphics[width=0.24\linewidth, clip, trim={#4}]{results/cmp_360loc/#2/#3/1.jpg} &
    \includegraphics[width=0.24\linewidth, clip, trim={#4}]{results/cmp_360loc/#2/#3/2.jpg} &
    \includegraphics[width=0.24\linewidth, clip, trim={#4}]{results/cmp_360loc/#2/#3/3.jpg}
    \\
}

\newcommand{\cmpframes}[3]{
    \begin{figure*}[t]
        \centering
        \small
        \setlength\tabcolsep{1px}
        \resizebox{1.\linewidth}{!}{
            \begin{tabular}{rcccc}
                \row{MVSplat}{mvsplat}{#1}{#2}
                \row{PanSplat}{pansplat}{#1}{#2}
                \row{PanSplat + Deferred BL}{pansplat_defbl}{#1}{#2}
                \row{GT}{gt}{#1}{#2}
                & Frame \#0 & Frame \#1 & Frame \#2 & Frame \#3
            \end{tabular}
        }
        \caption{
            \textbf{Qualitative comparisons on 360Loc dataset.}
            We show zoomed-in regions of the generated images by MVSplat and PanSplat, with (PanSplat + Deferred BL) and without (PanSplat) deferred blending, compared to the ground truth (GT).
            The different columns represent different frames in the sequence, where Frame \#0 and Frame \#3 of GT are input views.
            We render the images across all four views to visualize different frame distances.
        }\label{fig:cmp_360loc_#3}
    \end{figure*}
}

\cmpframes{atrium-daytime_360_0-000501_000504}{542px 50px 100px 80px}{1}
\cmpframes{atrium-daytime_360_1-000050_000053}{742px 0px 0px 230px}{2}
\cmpframes{atrium-nighttime_360_1-000093_000096}{742px 50px 0px 180px}{3}
\cmpframes{atrium-nighttime_360_1-000300_000303}{459px 55px 283px 175px}{4}

}
{

\newcommand{\row}[4]{
    \raisebox{11ex}{\rotatebox[origin=c]{90}{#1}} &
    \includegraphics[width=0.19\linewidth, clip, trim={#4}]{results/cmp_insta360/#2/#3/0.jpg} &
    \includegraphics[width=0.19\linewidth, clip, trim={#4}]{results/cmp_insta360/#2/#3/4.jpg} &
    \includegraphics[width=0.19\linewidth, clip, trim={#4}]{results/cmp_insta360/#2/#3/8.jpg} &
    \includegraphics[width=0.19\linewidth, clip, trim={#4}]{results/cmp_insta360/#2/#3/12.jpg} &
    \includegraphics[width=0.19\linewidth, clip, trim={#4}]{results/cmp_insta360/#2/#3/16.jpg}
    \\
}

\newcommand{\cmpframes}[3]{
    \begin{figure*}[t]
        \centering
        \small
        \setlength\tabcolsep{1px}
        \resizebox{1.\linewidth}{!}{
            \begin{tabular}{rccccc}
                \row{MVSplat}{mvsplat}{#1}{#2}
                \row{PanSplat}{pansplat}{#1}{#2}
                \row{PanSplat + Deferred BL}{pansplat_defbl}{#1}{#2}
                \row{GT}{gt}{#1}{#2}
                & Frame \#0 & Frame \#4 & Frame \#8 & Frame \#12 & Frame \#16
            \end{tabular}
        }
        \vspace{-0.5em}
        \caption{
            \textbf{Qualitative comparisons on Insta360 dataset.}
            We show zoomed-in regions of the generated images by MVSplat and PanSplat, with (PanSplat + Deferred BL) and without (PanSplat) deferred blending, compared to the ground truth (GT).
            The different columns represent different frames in the sequence, where Frame \#0 and Frame \#16 of GT are input views.
            We render the images across five evenly-spaced intermediate views to visualize different frame distances.
        }\label{fig:cmp_insta360_#3}
    \end{figure*}
}

\cmpframes{VID_20240914_103257_00_005-008839_008855}{703px 140px 119px 170px}{1}
\cmpframes{VID_20240914_103257_00_005-009712_009728}{109px 15px 433px 15px}{2}
\cmpframes{VID_20240922_102141_00_006-003972_003988}{509px 15px 133px 115px}{3}
\cmpframes{VID_20240922_102141_00_006-005959_005975}{409px 15px 333px 215px}{4}

}
\stepcounter{section}

In \cref*{sec:ablation} of the main paper, we evaluate how two-step deferred backpropagation saves memory consumption during training.
Here, we provide additional details on the both training and inference memory usage in~\cref{fig:gpu_memory_ab}.

\noindent\textbf{How do Fibo and 3DGP help save memory?}
Comparing PanSplat (Full) with ablated versions (w/o Fibo, w/o 3DGP), we find that although the removal of 3D Gaussian pyramid (w/o 3DGP) introduces less Gaussians, it still consumes more memory due to slightly larger memory footprint of single cost volume.
On the other hand, during inference, the removal of Fibonacci Gaussians (w/o Fibo) causes out-of-memory error starting from $1792 \times 3584$ resolution, a resolution that PanSplat can still support.

\noindent\textbf{How does deferred backpropagation help save memory?}
We then add deferred backpropagation (w/ Deferred BP) with tile settings of $2 \times 2$ (4 tiles) and $4 \times 4$ (16 tiles).
As shown, the memory consumption drops significantly, with 16 tiles further enabling 4K inference on a 24GB RTX 3090 GPU.
We use 4 tiles for fine-tuning at $1024 \times 2048$ resolution and 16 tiles for fine-tuning at $2048 \times 4096$ resolution, with a batch size of 3 and 1, respectively.

\noindent\textbf{How does two-step design based on cubemap renderer help save memory?}
We also include an ablated version with only step 2 of deferred backpropagation (1 step) with 16 tiles setting.
The results show that the one-step version consumes significantly more memory than the two-step version when training, showing the effectiveness of cubemap renderer in reducing memory consumption.
We note that the inference memory usage stays consistent as they share the same cubemap renderer with sequential face rendering.

\section{Demo Video}\label{sec:demo}

By enabling 4K resolution support, PanSplat becomes a promising solution for immersive VR and virtual tours applications.
We provide a demo video to demonstrate the superior image quality of PanSplat on diverse datasets, and to showcase its potential applications in real-world scenarios.

\end{document}